\documentclass{article}
\newcommand{\ArxivVersion}{}

\usepackage{PRIMEarxiv}
\usepackage{fancyhdr}

\usepackage{microtype}
\usepackage{graphicx}
\usepackage{subcaption}
\usepackage{booktabs} \usepackage[normalem]{ulem} 
\usepackage{hyperref}

\usepackage{algorithmic}

\usepackage{amsmath}
\usepackage{amssymb}
\usepackage{mathtools}
\usepackage{amsthm}

\usepackage[capitalize,noabbrev]{cleveref}

\theoremstyle{plain}

\theoremstyle{definition}

\theoremstyle{remark}

\usepackage{url}
\usepackage{multirow}
\usepackage{sidecap}

\usepackage{natbib}

\usepackage{amsmath,amsfonts,amssymb}
\usepackage{hyperref}
\usepackage{url}
\usepackage{float}
\usepackage{booktabs}
\usepackage{multirow}
\usepackage{graphicx}
\usepackage{subcaption}
\usepackage[normalem]{ulem}

\newcommand{\resizeTabular}[2][]{    \begingroup
    \if\relax\detokenize{#1}\relax
        \resizebox{\linewidth}{!}{            \LARGE                                    \setlength{\tabcolsep}{4pt}            #2        }    \else
        \scalebox{#1}{            \LARGE
            \setlength{\tabcolsep}{4pt}            #2        }    \fi
    \endgroup
}

\usepackage{xcolor}
\usepackage{eso-pic}
\usepackage{zref-abspage}

\newif\ifPageLimitAuditActive
\PageLimitAuditActivefalse

\newcommand{\PageLimitAuditLimit}{0}

\newcommand{\PageLimitAudit}[1]{  \gdef\PageLimitAuditLimit{#1}  \global\PageLimitAuditActivetrue
}

\AddToShipoutPictureBG{  \ifPageLimitAuditActive
    \ifnum\value{abspage}>\PageLimitAuditLimit\relax
      \AtPageLowerLeft{        \color{yellow!25}        \rule{\paperwidth}{\paperheight}      }    \fi
  \fi
}

\newcommand{\PageLimitAuditEnding}{  \ifPageLimitAuditActive
        \clearpage
    \ifnum\value{abspage}>\PageLimitAuditLimit\relax
      \PackageWarningNoLine{page-limit-audit}{        Audited content occupies \arabic{abspage} pages; limit is
        \PageLimitAuditLimit
      }    \fi
    \global\PageLimitAuditActivefalse
  \fi
}

\begin{document}

\renewcommand{\citep}[1]{\cite{#1}}
\renewcommand{\citet}[1]{Ref~\cite{#1}}

\ifdefined\ArxivVersion
    \else
    \PageLimitAudit{9} \fi

\title{MechBench: Can AI Scientific Agents Discover Mechanisms Beyond Phenomenal Laws?}
\author{Zihan Yu\textsuperscript{1}\thanks{Equal contribution.},
Jiadong Zhang\textsuperscript{2,3}\footnotemark[1],
Jialin Cheng\textsuperscript{4}\footnotemark[1],
Jingtao Ding\textsuperscript{5}\thanks{Corresponding authors.},
Yong Li\textsuperscript{1}\footnotemark[2] \\
\textsuperscript{1}Department of Electronic Engineering, BNRist, Tsinghua University, Beijing, China \\
\textsuperscript{2}Institute of Automation, Chinese Academy of Sciences, Beijing, China \\
\textsuperscript{3}Beijing Zhongguancun Academy, Beijing, China \\
\textsuperscript{4}Xi'an Jiaotong University, Xi'an, China \\
\textsuperscript{5}Department of Earth System Science, Tsinghua University, Beijing, China \\
\texttt{yuzh23@mails.tsinghua.edu.cn, \{dingjingtao, liyong07\}@tsinghua.edu.cn}
}
\maketitle

\begin{abstract}
Scientific discovery requires not only recovering mathematical laws that describe observable behavior, but also identifying the mechanisms that generate them. Existing benchmarks for symbolic regression and scientific agents primarily evaluate phenomenal-law recovery, leaving mechanism discovery largely untested. We introduce \textsc{MechBench}, a benchmark that explicitly separates these two capabilities. Each task is defined by a mechanistic model, a structured set of scientifically meaningful relations whose joint consequences entail an observable phenomenal law, while agents receive only observational data and scientific context. We evaluate mechanism recovery through \emph{mechanism probes}, which query internal scientific consequences that cannot be inferred from the phenomenal law alone. To reduce reliance on memorized textbook mechanisms, we construct unfamiliar variants through controlled, scientifically interpretable mutations of canonical mechanisms, and screen for \emph{mechanistic indistinguishability} to exclude ambiguous instances admitting comparable competing mechanisms. Experiments across representative scientific agents reveal a substantial phenomenal--mechanism recovery gap: for Codex with GPT-5.6-sol, phenomenal-law accuracy reaches $35.00\%$ on the Core-set while mechanism accuracy is only $13.75\%$, with mechanism recovery failing in $64.29\%$ of cases where the phenomenal law is correctly recovered. The gap widens as mechanisms become increasingly mutated, and even providing the correct phenomenal law leaves mechanism recovery below $50\%$. These results reveal a substantial generalization gap in mechanistic reasoning and establish mechanism discovery as a distinct challenge beyond recovering observable scientific laws.
\end{abstract}

\section{Introduction}
\label{sec:introduction}

Discovering compact mathematical laws from observations has long been a central objective of science and, more recently, of automated scientific discovery. Yet recovering a law that describes observable behavior is not the same as recovering the mechanism that produces it. In the philosophy of science, \textit{phenomenal models} characterize the observable inputs, modulators, and outputs of a system while leaving its relevant internal causal structure unspecified, whereas \textit{mechanistic models} further describe the components, relations, and organization responsible for generating the phenomenon \citep{Machamer2000,KaplanCraver2011,CraverKaplan2020}. The distinction is exemplified by the transition from Kepler's empirical laws of planetary motion to Newtonian mechanics: the former compactly characterize regularities among observable quantities, while the latter provides an underlying structure from which those regularities can be derived. Mechanistic models therefore make commitments beyond observational adequacy: they seek to explain how a phenomenon is generated and imply properties of scientifically meaningful internal quantities that are not specified by the observable law itself. Assessing whether AI systems can recover such mechanisms is thus a distinct and stronger test of scientific discovery than accurate equation recovery.

Large language models and scientific agents have recently shown rapidly improving capabilities for discovering mathematical regularities from data, progressing from LLM-guided symbolic regression to autonomous systems that analyze observations, conduct experiments, invoke scientific tools, and iteratively refine hypotheses \citep{Shojaee2025LLMSR,Xia2026SRScientist,Yang2026KeplerAgent}. Accordingly, recent benchmarks have substantially strengthened the evaluation of scientific law discovery: they introduce unfamiliar or counterfactual laws to reduce memorization, allow agents to actively interrogate simulated systems, and increasingly ground discovery in realistic scientific contexts and data \citep{Shojaee2025LLMSRBench,Zheng2026NewtonBench,Cerrato2026ScienceGym,Huang2026SciLawsBench}. Despite these advances, their principal target remains the recovery of mathematical relations that reproduce observable system behavior. Such evaluations establish whether an agent can discover an observationally adequate law, but do not require it to reconstruct any internal mechanism from which that law follows. A recent work argues that a central missing capability in current LLMs is the abductive ``jump'' from observations to explanatory premises, distinguishing it from induction over observed patterns and deduction from given premises \citep{zahavyposition}. This perspective highlights a capability that existing law-discovery benchmarks do not directly test: whether AI systems can infer explanatory internal structure beyond observable regularities.

Extending evaluation from law discovery to mechanism discovery is nontrivial. A benchmark must specify mechanistic structure in a form that can be evaluated objectively, distinguish genuine reconstruction from recall of familiar textbook mechanisms, and account for \emph{mechanistic indistinguishability}, where distinct scientifically plausible mechanisms entail the same observable law. These challenges make it difficult to determine, under controlled conditions, whether the growing law-discovery capability of scientific agents extends to recovering underlying mechanisms.

\begin{figure*}[t]
    \centering
    \includegraphics[width=\textwidth]{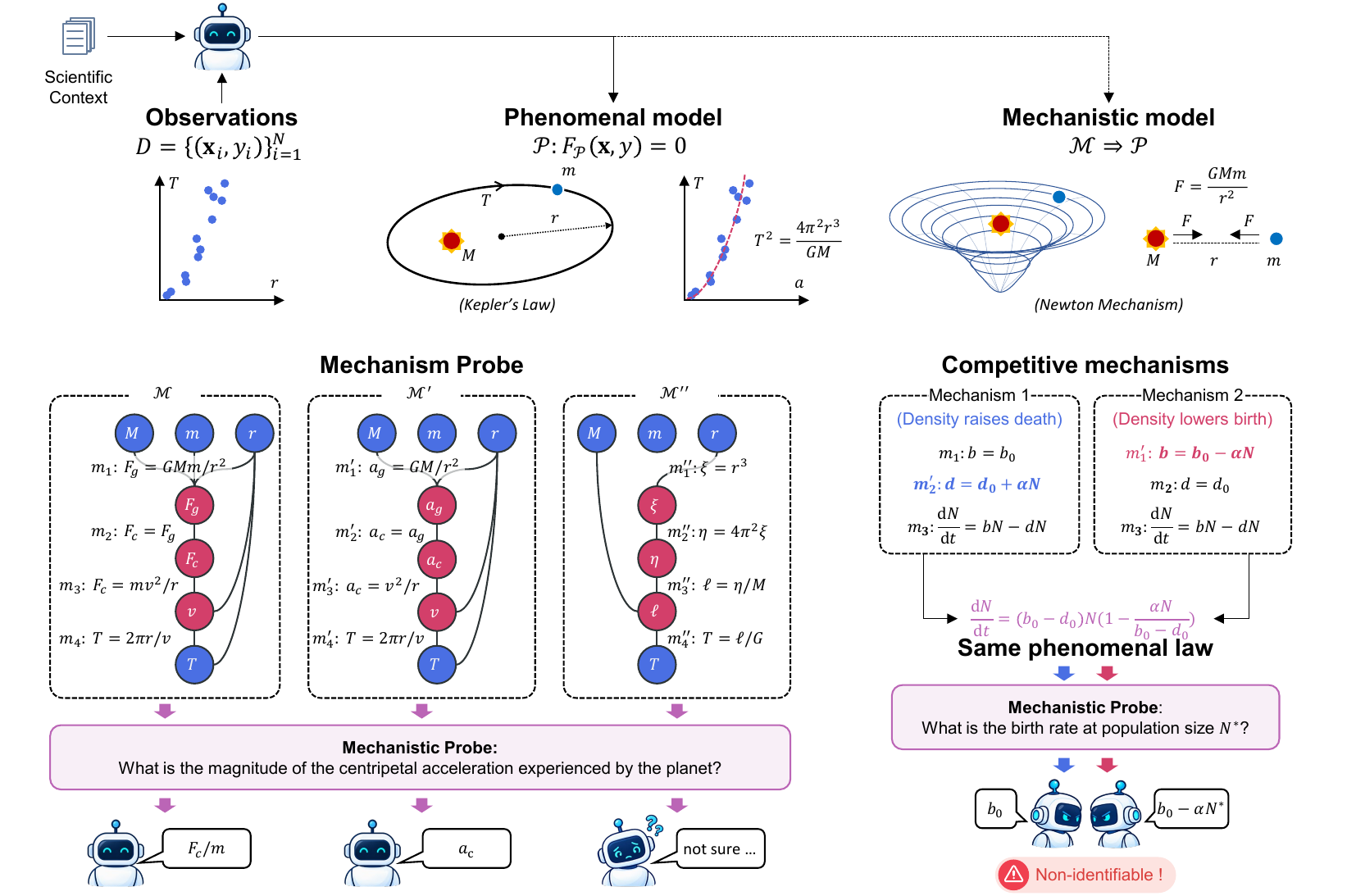}
    \caption{From phenomenal-law recovery to mechanistic discovery.}
    \label{fig:main-idea}
\end{figure*}

To address these challenges, we introduce \textsc{MechBench}, a benchmark that explicitly separates the recovery of observable laws from the recovery of the mechanisms that generate them, as illustrated in Figure~\ref{fig:main-idea}. We represent a mechanism as a structured set of scientifically meaningful relations whose joint consequences entail an observable phenomenal law, allowing an agent to be evaluated not only on whether its proposed mechanism reproduces the observed input--output relationship, but also on whether it makes the correct additional commitments about internal scientific quantities. We operationalize the latter through \emph{mechanism probes}, which query scientifically defined internal consequences that follow from the submitted mechanism but cannot be inferred from the phenomenal law and provided scientific context alone. To distinguish reconstruction from memorization, we construct unfamiliar mechanism variants through controlled, scientifically interpretable mutations of canonical mechanisms. Finally, we explicitly screen for mechanistic indistinguishability and revise or exclude instances admitting readily constructible competing mechanisms with the same phenomenal law. Together, these components turn mechanism discovery from an informal notion of ``deeper understanding'' into a controlled and objectively testable discovery task.

Experiments across representative scientific agents reveal a substantial gap between phenomenal-law and mechanism recovery. Taking Codex with GPT-5.6-sol as an example, phenomenal-law accuracy reaches $35.00\%$ on the Core-set while mechanism accuracy is only $13.75\%$, and mechanism recovery still fails in $64.29\%$ of cases where the phenomenal law is recovered correctly. This separation grows as mechanisms depart further from their canonical forms: phenomenal-law and mechanism recovery decline together with increasing mutation count, but mechanism recovery deteriorates more sharply. Moreover, even when the correct phenomenal law is directly provided, mechanism accuracy remains below $50\%$ and drops rapidly on increasingly mutated mechanisms. These results expose a substantial generalization gap in mechanistic reasoning and suggest that mechanism discovery constitutes an additional, qualitatively more difficult challenge beyond observable-law recovery.

\section{Related Work}

\textbf{LLMs for Scientific Equation Discovery.}
Recent LLM-based symbolic-regression and scientific-agent systems increasingly combine scientific prior knowledge with program synthesis, data analysis, tool use, and iterative hypothesis refinement. LLM-SR integrates language-model proposals with evolutionary search \citep{Shojaee2025LLMSR}; SR-Scientist equips agents with data-analysis and equation-evaluation tools for iterative discovery \citep{Xia2026SRScientist}; and systems such as KeplerAgent and A-SR further incorporate structural reasoning, tool orchestration, and longer agentic search trajectories \citep{Yang2026KeplerAgent,Zhao2026ASR}. These approaches substantially extend symbolic regression beyond blind expression search, but their primary target remains the recovery of mathematical relations that characterize observable system behavior. Our setting instead evaluates whether an agent can recover the scientific internal relations from which those observable relations follow.

\textbf{Benchmarks for Scientific Law Discovery.}
Classical symbolic-regression benchmarks, including the Feynman dataset and SRBench, evaluate the recovery of compact equations from predefined observations \citep{UdrescuTegmark2020,LaCava2021SRBench,Matsubara2024SRSD}. More recent benchmarks strengthen this setting for LLM-based scientific discovery: LLM-SRBench uses transformed and synthetic equations to reduce direct memorization \citep{Shojaee2025LLMSRBench}; NewtonBench places altered physical laws in interactive simulated systems \citep{Zheng2026NewtonBench}; Science-Gym evaluates autonomous experimentation and data collection \citep{Cerrato2026ScienceGym}; and SciLaws-Bench combines real scientific observations with controlled, actively queryable scientific worlds \citep{Huang2026SciLawsBench}. Together, these benchmarks increase unfamiliarity, interactivity, and scientific grounding, while their primary object of evaluation remains the recovered observable law. MechBench instead targets mechanism discovery: it evaluates whether agents can recover structured mechanistic models whose internal scientific consequences agree with those of the reference mechanism, using controlled variants, mechanism probes, and screening for mechanistic indistinguishability.

\section{From Phenomenal Laws to Mechanistic Discovery}

\subsection{Formalizing Mechanistic Discovery}
\label{sec:formalizing-mechanistic-discovery}

Let $\mathbf{x}$ denote a set of observable input variables and $y$ the target variable. A phenomenal law $\mathcal{P}$ of the form $F_{\mathcal P}(\mathbf{x},y)=0$ represents their observable relationship. A mechanism $\mathcal{M}=\{m_1,\ldots,m_J\}$, on the other hand, consists of a collection of scientifically meaningful relations involving the observable variables and internal quantities $\mathbf{h}$, where $\mathbf{h}$ denotes quantities that cannot be directly observed. By eliminating $\mathbf{h}$ from $\mathcal{M}$, we can derive the phenomenal law $\mathcal{P}$:
\begin{equation}
\mathcal{M}(\mathbf{x},y,\mathbf{h}) \Longrightarrow \mathcal{P}(\mathbf{x},y).
\label{eq:mechanism-phenomenon}
\end{equation}
The additional internal structure encoded by $\mathcal{M}$ allows a mechanistic model to determine scientifically meaningful quantities that cannot be inferred from $\mathcal{P}$ alone. For example, the phenomenal relation for Kepler's third law can be written as $T^2=4\pi^2r^3/(GM)$. A Newtonian mechanism instead specifies gravity, orbital acceleration, force balance, and velocity--period coupling.

In this work, we formulate mechanistic discovery as the task of reconstructing a mechanistic model from observational data and scientific context. A reference mechanism $\mathcal{M}$ induces a phenomenal law $\mathcal{P}$, from which observations $D=\{(\mathbf{x}_i,y_i)\}_{i=1}^{N}$ can be sampled. As demonstrated in Figure~\ref{fig:main-idea}, the agent receives $D$ and scientific context $C$ that specifies the descriptions of observable variables and other background information, and is asked to reconstruct the underlying mechanistic model without access to the reference mechanism or its internal quantities. Mechanistic discovery is therefore formulated as
\begin{equation}
(C,D)\longrightarrow\hat{\mathcal M}.
\label{eq:mechanism-discovery}
\end{equation}

\subsection{Mechanism Probes}
\label{sec:mechanism-probes}

The same mechanism may be represented in different but scientifically equivalent ways. For example, as illustrated in Figure~\ref{fig:main-idea}, one formulation may explicitly reason through force, whereas another may eliminate force and express the same physical relations in terms of acceleration. Mechanism recovery therefore cannot be assessed through literal agreement with the equations used to represent the reference mechanism, as exact matching of equations, intermediate symbols, or dependency graphs would conflate representational and mechanistic differences. On the other hand, evaluating a submitted mechanism only by eliminating its internal quantities and comparing the resulting phenomenal law with the reference $\mathcal{P}$ is also insufficient, because scientifically distinct mechanisms may entail the same phenomenal law.

Inspired by the mechanistic view that explaining a phenomenon requires specifying scientifically meaningful internal relations that generate the observable regularity and support correct reasoning about internal quantities that are not determined by the phenomenal law itself, we introduce \emph{mechanism probes} to assess these additional mechanistic commitments. Specifically, we select a set $\mathcal Z=\{z_k\}_{k=1}^{K}$ of scientifically defined quantities such that, for a generic probe $z\in\mathcal Z$,
\begin{equation}
\mathcal{M}\Rightarrow z=g_{\mathcal M}(\mathbf{x}), \qquad
\mathcal{P},C\not\Rightarrow z=g_{\mathcal M}(\mathbf{x}).
\label{eq:mechanism-probe}
\end{equation}
The first condition requires each probe to capture an internal consequence of the reference mechanism, while the second ensures that it cannot be inferred from the phenomenal law and the provided scientific context alone. A properly chosen probe can then be used to assess whether $\hat{\mathcal M}$ reproduces the scientifically equivalent mechanistic commitments. Given its submitted mechanism, the agent is asked to derive
\begin{equation}
\hat{\mathcal M}\Rightarrow z=g_{\hat{\mathcal M}}(\mathbf{x}),
\label{eq:submitted-mechanism-probe}
\end{equation}
which can then be compared with the reference probe relation $z=g_{\mathcal M}(\mathbf{x})$ symbolically and numerically. Because each probe is defined by the scientific meaning of the queried quantity rather than by a particular intermediate symbol, equivalent formulations of the same mechanism should imply the same probe value even when their derivations differ. In practice, we restrict probes to quantities with clear scientific interpretations and explicit, objectively checkable expressions from the variables available in the task; Appendix~\ref{app:probe-construction} details their construction.

\subsection{Mechanistic Indistinguishability}
\label{sec:mechanistic-indistinguishability}

Since eliminating the internal quantities $\mathbf{h}$ from $\mathcal M$ yields the phenomenal law $\mathcal P$, $\mathcal M$ can be viewed algebraically as a decomposition of $\mathcal P$. However, not every mathematically valid decomposition constitutes a mechanistic model. The provided scientific context $C$ strongly constrains which entities, relations, and assumptions provide a meaningful explanation of the phenomenon. This constraint makes mechanism discovery well posed despite the mathematical underdetermination.

However, the problem becomes ambiguous when there exists a competing mechanism $\mathcal M'$ that is substantively different from the reference mechanism $\mathcal M$ but entails the same phenomenal law and remains comparably admissible under the provided scientific context $C$. As illustrated in Figure~\ref{fig:main-idea}, the same population-level law may arise either from density-dependent suppression of birth or from density-dependent enhancement of death, while the two mechanisms make different predictions about the underlying birth rate. Such cases are a mechanism-level form of empirical equivalence and underdetermination by evidence \citep{HoeferRosenberg1994}, which can be referred to as \emph{mechanistic indistinguishability}. In this case, the observable evidence together with the provided scientific context offers no sufficient basis for singling out $\mathcal M$ over $\mathcal M'$, making recovery of the designated reference mechanism intrinsically ambiguous. Avoiding such ambiguity is also necessary for interpreting the mechanism probes introduced in Section~\ref{sec:mechanism-probes}. A mechanism probe tests whether a proposed mechanism $\hat{\mathcal M}$ reproduces the same scientifically meaningful internal consequences as the reference mechanism $\mathcal M$, but does not determine which of two competing mechanisms is scientifically more plausible. Probe-based agreement or disagreement therefore provides a valid operational criterion for mechanism recovery only when no substantively different mechanism that is comparably admissible under $C$ remains compatible with the same phenomenal evidence. We therefore require benchmark instances to avoid such competing mechanisms; the corresponding screening procedure is described in Section~\ref{sec:constructing-benchmark-instances}.

\section{Benchmark Construction}
\label{sec:benchmark-construction}

\begin{figure*}
    \centering
    \includegraphics[width=\textwidth]{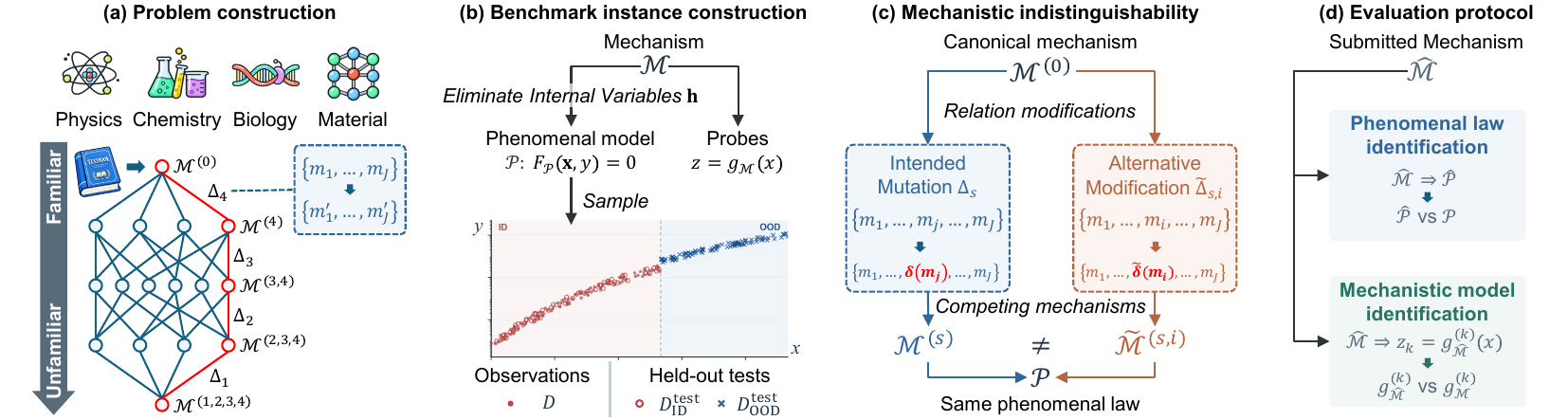}
    \caption{Construction and evaluation of MechBench. (a) Canonical mechanisms are modified and composed to create unfamiliar variants. (b) Each mechanism induces a phenomenal law used to generate observations and held-out ID/OOD tests. (c) Candidate competing mechanisms are screened for mechanistic indistinguishability. (d) Submitted mechanisms are evaluated by both phenomenal-law recovery and mechanism probes.}
    \label{fig:mechbench}
\end{figure*}

\subsection{Constructing Mechanism Variants}
\label{sec:constructing-mechanism-variants}

Scientific-discovery benchmarks such as AI Feynman and LLM-SRBench often construct tasks from established scientific laws and equations \citep{UdrescuTegmark2020,Shojaee2025LLMSRBench}. For LLM-based agents, however, canonical scientific knowledge may already be present in pretraining data, making successful recovery difficult to distinguish from memorization \citep{Shojaee2025LLMSRBench}. To balance scientific grounding with mechanism unfamiliarity, we select canonical scientific mechanisms from physics, chemistry, biology, and materials science and use them as blueprints for constructing unfamiliar mechanism variants (Appendix~\ref{app:mechanism-families}).

Specifically, a mutation $\Delta_s$ applies a scientifically interpretable modification to the canonical mechanism,
\begin{equation}
\mathcal M^{(s)} = \Delta_s\!\left(\mathcal M^{(0)}\right).
\label{eq:mechanism-mutation}
\end{equation}
A useful mutation should depart sufficiently from the canonical mechanism to prevent straightforward reproduction of its textbook derivation, while remaining scientifically admissible under the relevant domain knowledge. We therefore favor local modifications with clear scientific semantics over arbitrary symbolic perturbations. Representative mutations include changing an interaction scaling law, introducing a source or leakage term into a balance relation, replacing a linear constitutive response with a nonlinear one, or modifying a geometric constraint; additional examples and details are provided in Appendix~\ref{app:mechanism-mutations}.

Compatible mutations can further be composed to construct variants that depart more substantially from the canonical mechanism and are less amenable to direct textbook recall. As illustrated in Figure~\ref{fig:mechbench}(a), given a set of compatible mutations $\{\Delta_{s_1},\ldots,\Delta_{s_q}\}$, we construct
\begin{equation}
\mathcal M^{(s_1,\ldots,s_q)} = (\Delta_{s_q}\circ\cdots\circ\Delta_{s_1}) \left(\mathcal M^{(0)}\right).
\label{eq:mechanism-composition}
\end{equation}
The five mutations in each family are selected and, where needed, adjusted to remain compatible under composition. The resulting task family retains a shared scientific scaffold while varying which and how many elements of the canonical mechanism are modified, from the original through all mutation subsets.

\subsection{Constructing Benchmark Instances}
\label{sec:constructing-benchmark-instances}

For each constructed mechanism $\mathcal M$, we derive its corresponding phenomenal law $\mathcal P$ by eliminating the internal quantities $\mathbf h$ and use this law to generate observable data. As illustrated in Figure~\ref{fig:mechbench}(b), let $\mathcal{X}$ denote the scientifically meaningful range of the observable inputs. We partition this range into two disjoint regions, $\mathcal X_{\mathrm{ID}}$ and $\mathcal X_{\mathrm{OOD}}$, corresponding respectively to interpolation and extrapolation regimes. We sample observations $D$ from $\mathcal X_{\mathrm{ID}}$ to provide to the agent, and we construct held-out test sets
\begin{equation}
D_{\mathrm{ID}}^{\mathrm{test}} =\{(\mathbf{x}_i^{\mathrm{ID}},y_i^{\mathrm{ID}})\}_{i=1}^{N_{\mathrm{ID}}},\quad \mathbf{x}_i^{\mathrm{ID}}\in\mathcal X_{\mathrm{ID}},
\qquad
D_{\mathrm{OOD}}^{\mathrm{test}} =\{(\mathbf{x}_i^{\mathrm{OOD}},y_i^{\mathrm{OOD}})\}_{i=1}^{N_{\mathrm{OOD}}},\quad \mathbf{x}_i^{\mathrm{OOD}}\in\mathcal X_{\mathrm{OOD}},
\label{eq:id-ood-data}
\end{equation}
to evaluate in-domain and out-of-domain numerical consistency of the recovered phenomenal law. Appendix~\ref{app:observation-generation} describes the sampling ranges, public task view, and hidden evaluation data.

To avoid constructing mechanistically indistinguishable instances, we explicitly search for competing mechanisms during instance construction. As illustrated in Figure~\ref{fig:mechbench}(c), consider a canonical mechanism $\mathcal M^{(0)}=\{m_1,\ldots,m_J\}$ and a variant $\mathcal M^{(s)}=\Delta_s(\mathcal M^{(0)})$ in which the intended mutation replaces relation $m_j$ with $\delta(m_j)$, so that $\mathcal M^{(s)}=\{m_1,\ldots,\delta(m_j),\ldots,m_J\}$. We derive candidate alternative modifications $\tilde{\delta}(m_i)$ of other relations $m_i$ ($i\neq j$) that can reproduce the same induced phenomenal law, creating
\begin{equation}
\widetilde{\mathcal M}^{(s,i)} =\{m_1,\ldots,\tilde{\delta}(m_i),\ldots,m_J\}.
\end{equation}
If such an alternative requires comparable or lower structural complexity and no greater departure from the relevant scientific background knowledge than the intended modification $\delta(m_j)$, this provides strong evidence that the resulting task is mechanistically indistinguishable, and the corresponding variant is revised or excluded.

\subsection{Benchmark Evaluation Protocol}
\label{sec:evaluation-protocol}

As illustrated in Figure~\ref{fig:mechbench}(d), for a submitted mechanism $\hat{\mathcal M}$, our evaluation separates two complementary signals: whether $\hat{\mathcal M}$ recovers the phenomenal law and whether it reproduces the additional mechanistic consequences encoded by the mechanism probes. To evaluate phenomenal-law recovery, we first derive the phenomenal law implied by the submission,
\begin{equation}
\hat{\mathcal M}\Longrightarrow\hat{\mathcal P},
\label{eq:evaluate-phenomenal-law}
\end{equation}
and compare $\hat{\mathcal P}$ with the reference $\mathcal P$ using symbolic and numerical equivalence. This determines whether the proposed mechanism accounts for the observed phenomenal relationship.

To evaluate mechanism recovery, we freeze the submitted mechanism and present the scientific definition of each mechanism probe $z_k$ to the agent independently, asking it to derive the relation between the defined quantity and the observable variables based on its submitted $\hat{\mathcal M}$,
\begin{equation}
\hat{\mathcal M}
\Longrightarrow
z_k=g_{\hat{\mathcal M}}^{(k)}(\mathbf{x}).
\label{eq:evaluate-mechanism-probe}
\end{equation}
The predicted relation $g_{\hat{\mathcal M}}^{(k)}$ is compared with the reference relation $g_{\mathcal M}^{(k)}$ using the same symbolic and numerical equivalence machinery. Probes are queried independently so that information revealed by one probe cannot alter the mechanism used to answer another. 

\section{Experiments}
\label{sec:experiments}

\subsection{Experimental Setup}
\label{sec:experimental-setup}

\paragraph{Benchmark and evaluated systems.} We evaluate on two MechBench subsets constructed in Section~\ref{sec:benchmark-construction} and detailed in Appendix~\ref{app:task-construction-details}. The \emph{Full-set} contains 512 instances from 16 task families spanning physics, chemistry, biology, and materials science, while the \emph{Core-set} is a fixed subset of 80 instances from eight families. We use the Core-set to compare a broader range of model--agent configurations and the Full-set to evaluate performance across more task families. We evaluate seven configurations spanning Codex, Claude Code, and DeepSeek Harness with GPT-5.6-sol, GLM-5.3-flash, DeepSeek-v4-flash-0731, and DeepSeek-v4-pro-0813. All seven configurations are evaluated on the Core-set, while Codex with GPT-5.6-sol and GLM-5.3-flash is additionally evaluated on the Full-set. We also include PySR as a non-LLM equation-discovery baseline and PySR + Direct-Ask, where DeepSeek-v4-flash-0731 answers mechanism probes from the equation discovered by PySR.

\paragraph{Gold-$P$ control.} To isolate mechanism reasoning from phenomenal-law discovery, we additionally evaluate \emph{Gold-$P$ Direct-Ask}, where the reference phenomenal law $P$ is provided directly and the model answers each mechanism probe independently in a tool-free turn. We evaluate this control with GPT-5.6-sol, GLM-5.3-flash, and DeepSeek-v4-flash-0731 on both the Core-set and Full-set.

\paragraph{Protocol and metrics.} In the standard discovery setting, agents receive only the scientific context $C$ and training observations $D$; the reference mechanism, mechanism probes, and held-out observations remain inaccessible during search. We report phenomenal-law accuracy $\mathrm{SA}(P)$ and task-level mechanism accuracy $\mathrm{SA}(M)$, where a task is counted as mechanistically correct only if all of its probes are symbolically correct. Detailed configurations, numerical ID/OOD metrics, scoring rules, and repeated-run aggregation are provided in Appendix~\ref{app:experimental-details}.

\subsection{The Phenomenal--Mechanism Recovery Gap}
\label{sec:recovery-gap}

Table~\ref{tab:main-recovery} reports phenomenal-law and mechanism recovery in the standard discovery setting. Across all seven agent configurations, mechanism accuracy is consistently lower than phenomenal-law accuracy, revealing a systematic gap between recovering observable behavior and reconstructing the mechanism that generates it. Taking Codex with GPT-5.6-sol as an example, $\mathrm{SA}(P)$ reaches $35.00\%$ on the Core-set, whereas $\mathrm{SA}(M)$ is only $13.75\%$; on the Full-set, the corresponding scores are $16.02\%$ and $7.81\%$. The same pattern holds across the remaining configurations despite their substantially lower overall recovery rates. (See Appendix~\ref{app:variant-results} for variant-only results.)

To separate mechanism failures from failures to recover the observable law, we additionally report the conditional failure rate $1-(M\mid P)$, which is the probability of failing mechanism recovery among instances with successful phenomenal-law recovery (i.e., $\Pr(S_M=0 \mid S_P=1)$). For Codex with GPT-5.6-sol, this rate is $64.29\%$ on the Core-set and $57.32\%$ on the Full-set, suggesting that even when the phenomenal law is recovered correctly, mechanism recovery fails in more than half of the cases. The other agent configurations exhibit the same pattern, with Core-set conditional failure rates ranging from $49.07\%$ to $67.22\%$. These results suggest that mechanism recovery is not simply a stricter version of equation recovery: agents that successfully identify observable relations still frequently fail to reconstruct the internal scientific structure that produces them.

Moreover, Figure~\ref{fig:recovery-by-mutations} examines this rate as a function of mutation count for Codex with GPT-5.6-sol. Although phenomenal-law and mechanism accuracy decline together as mutations accumulate, mechanism accuracy deteriorates more sharply, causing the conditional mechanism failure rate to rise. Specifically, the rate is approximately $33\%$ on the Core-set and $17\%$ on the Full-set for Original instances, and increases rapidly with additional mutations, reaching $100\%$ at three mutations on the Core-set and four mutations on the Full-set. This widening separation suggests that increasing mechanism unfamiliarity disproportionately impairs recovery of internal mechanistic structure relative to recovery of the observable law, indicating that mechanism discovery poses an additional and qualitatively more difficult challenge than phenomenal-law discovery.

\begin{table*}[t]
    \centering
    \caption{Symbolic recovery results on the Core-set and Full-set. $\mathrm{SA}(P)$ and $\mathrm{SA}(M)$ denote phenomenal-law and mechanism recovery, respectively.}
    \label{tab:main-recovery}
    \resizeTabular{\begin{tabular}{ccccccccc}
\toprule
\multicolumn{1}{c}{\multirow{2}[4]{*}{\textbf{Base Model}}} & \multicolumn{1}{c}{\multirow{2}[4]{*}{\textbf{Agent}}} & \multicolumn{3}{c}{\textbf{Core-set  (n=80)}} &       & \multicolumn{3}{c}{\textbf{Full-set  (n=512)}} \\
\cmidrule{3-5}\cmidrule{7-9}      &       & \textbf{SA($\mathcal P$)↑} & \textbf{SA($\mathcal M$)↑} & \textbf{$1 - (\mathcal M\mid \mathcal P)$} &       & \textbf{SA($\mathcal P$)↑} & \textbf{SA($\mathcal M$)↑} & \textbf{$1-(\mathcal M\mid \mathcal P)$} \\
\midrule
Deepseek-v4-flash-0731 & PySR+Direct-Ask & \textbf{0.00\%} & \textbf{1.25\%} & /     &       & \textbf{0.20\%} & \textbf{0.20\%} & / \\
\midrule
GPT-5.6-sol & Codex & \textbf{35.00\%} & \textbf{13.75\%} & 64.29\% &       & \textbf{16.02\%} & \textbf{7.81\%} & 57.32\% \\
GLM-5.3-flash & Codex & 6.25\% & 2.92\% & 67.22\% &       & \uline{4.30\%} & \uline{1.76\%} & 63.64\% \\
GLM-5.3-flash & Claude Code & 8.75\% & \uline{3.75\%} & 57.14\% &       & /     & /     & / \\
Deepseek-v4-flash-0731 & Codex & 7.92\% & \uline{3.75\%} & 49.07\% &       & /     & /     & / \\
Deepseek-v4-flash-0731 & Claude Code & 6.25\% & \uline{3.75\%} & 60.00\% &       & /     & /     & / \\
Deepseek-v4-flash-0731 & Deepseek Harness & 6.25\% & 2.50\% & 60.00\% &       & /     & /     & / \\
Deepseek-v4-pro-0813 & Deepseek Harness & \uline{10.00\%} & \uline{3.75\%} & 62.50\% &       & /     & /     & / \\
\bottomrule
\end{tabular}}
\end{table*}

\subsection{Oracle Control: Knowing the Law Is Not Enough}
\label{sec:oracle-controls}

We use Gold-P Direct-Ask to test whether models can reliably answer mechanism probes when the correct phenomenal law is provided. According to Table~\ref{tab:oracle-controls}, agents achieve higher mechanism accuracy under Gold-P Direct-Ask than under standard discovery across both datasets for every model evaluated in both settings. For GPT-5.6-sol, accuracy reaches $45.00\%$ on the Core-set and $49.02\%$ on the Full-set, compared with $13.75\%$ and $7.81\%$, respectively, under standard discovery. Nevertheless, mechanism accuracy remains below $50\%$ on both datasets, while the two flash models remain substantially less successful. Thus, access to the correct observable relationship does not make mechanism recovery reliable. This is partly because the phenomenal law itself does not encode the internal mechanistic information required by the probes; on the other hand, it also reflects models' limited ability to reconstruct such missing mechanistic structure even when the observable law is given, underscoring the need for dedicated evaluation of mechanism discovery.

\begin{figure*}[t]
    \centering
    \begin{minipage}[t]{0.49\textwidth}
        \centering
        \captionof{figure}{Recovery accuracy and conditional mechanism failure rate across mutation counts for Codex with GPT-5.6-sol.}
        \label{fig:recovery-by-mutations}
        \includegraphics[width=\linewidth]{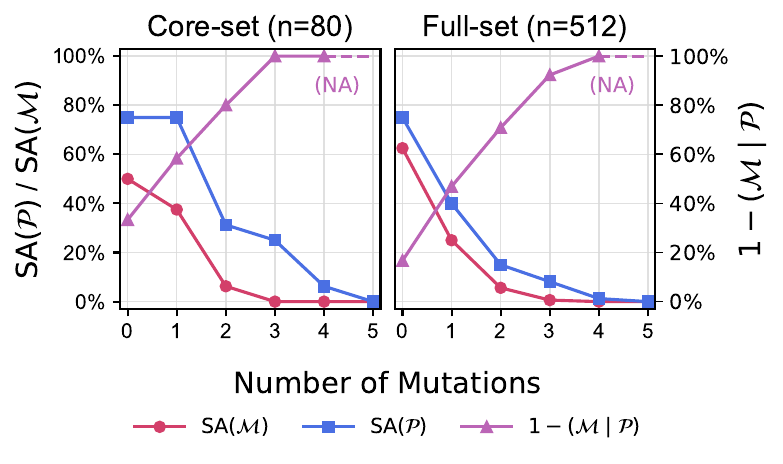}
    \end{minipage}\hfill
    \begin{minipage}[t]{0.49\textwidth}
        \centering
        \captionof{figure}{Mechanism recovery by mutation count under standard and Gold-P settings.}
        \label{fig:gold-p-by-mutations}
        \includegraphics[width=\linewidth]{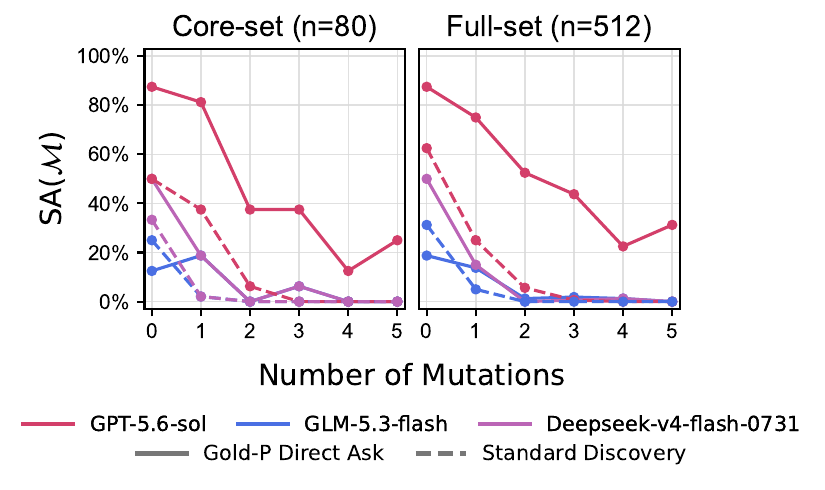}
    \end{minipage}
\end{figure*}

Figure~\ref{fig:gold-p-by-mutations} further shows that mechanism recovery under Gold-P deteriorates rapidly as mutation count increases. For GPT-5.6-sol, mechanism accuracy is $87.50\%$ on Original instances in both sets, but falls to $12.50\%$ on the Core-set and $22.50\%$ on the Full-set with four mutations; the two flash models remain near zero across most multi-mutation groups. The sharp degradation from Original to increasingly mutated mechanisms suggests a substantial generalization gap in mechanistic reasoning: strong performance on familiar scientific mechanisms does not readily transfer to novel mechanistic structures, highlighting the value of controlled mutations for evaluating mechanism discovery.

\begin{table*}[t]
    \centering
    \caption{Mechanism recovery under standard discovery and Gold-P Direct-Ask on the Core-set and Full-set.}
    \label{tab:oracle-controls}
    \resizeTabular{\begin{tabular}{lccccc}
\toprule
\multicolumn{1}{r}{\multirow{2}[4]{*}{\textbf{Base Model}}} & \multicolumn{2}{c}{\textbf{Core-set  (n=80) · SA(M)↑}} &       & \multicolumn{2}{c}{\textbf{Full-set  (n=512) · SA(M)↑}} \\
\cmidrule{2-3}\cmidrule{5-6}      & Standard discovery & Gold-P Direct-Ask &       & Standard discovery & Gold-P Direct-Ask \\
\midrule
GPT-5.6-sol & \textbf{13.75\%} & \textbf{45.00\%} &       & \textbf{7.81\%} & \textbf{49.02\%} \\
GLM-5.3-flash & 2.92\% & 6.25\% &       & 1.76\% & 3.91\% \\
Deepseek-v4-flash-0731 & \uline{3.75\%} & \uline{10.00\%} &       & /     & \uline{4.10\%} \\
\bottomrule
\end{tabular}}
\end{table*}

\subsection{Comparisons Across Models, Harnesses, and Task Families}
\label{sec:factor-comparisons}

\begin{table*}[b!]
    \centering
    \caption{Complete Core-set symbolic-recovery results by scientific domain. A dash denotes a metric that is not defined for the corresponding control or baseline.}
    \label{tab:appendix-core-domain-recovery}
    \resizeTabular{\begin{tabular}{llcccccccc}
\toprule
\multirow{2}{*}{\textbf{Base model}} & \multirow{2}{*}{\textbf{Agent / condition}}
& \multicolumn{2}{c}{\textbf{Physics}} & \multicolumn{2}{c}{\textbf{Chemistry}}
& \multicolumn{2}{c}{\textbf{Biology}} & \multicolumn{2}{c}{\textbf{Materials}} \\
\cmidrule(lr){3-4}\cmidrule(lr){5-6}\cmidrule(lr){7-8}\cmidrule(lr){9-10}
& & $\mathcal P$ & $\mathcal M$ & $\mathcal P$ & $\mathcal M$ & $\mathcal P$ & $\mathcal M$ & $\mathcal P$ & $\mathcal M$ \\
\midrule
No LLM & PySR & 0.00 & -- & 0.00 & -- & 0.00 & -- & 0.00 & -- \\
DeepSeek-v4-flash-0731 & PySR + Direct-Ask & -- & 5.00 & -- & 0.00 & -- & 0.00 & -- & 0.00 \\
\midrule
GPT-5.6-sol & Codex & 40.00 & 20.00 & 40.00 & 25.00 & 40.00 & 10.00 & 20.00 & 0.00 \\
GLM-5.3-flash & Codex (3-run avg.) & 11.67 & 11.67 & 5.00 & 0.00 & 3.33 & 0.00 & 5.00 & 0.00 \\
GLM-5.3-flash & Claude Code & 25.00 & 15.00 & 5.00 & 0.00 & 0.00 & 0.00 & 5.00 & 0.00 \\
DeepSeek-v4-flash-0731 & Codex (3-run avg.) & 15.00 & 10.00 & 8.33 & 5.00 & 3.33 & 0.00 & 5.00 & 0.00 \\
DeepSeek-v4-flash-0731 & Claude Code & 5.00 & 10.00 & 15.00 & 5.00 & 0.00 & 0.00 & 5.00 & 0.00 \\
DeepSeek-v4-flash-0731 & DeepSeek Harness & 15.00 & 10.00 & 0.00 & 0.00 & 5.00 & 0.00 & 5.00 & 0.00 \\
DeepSeek-v4-pro-0813 & DeepSeek Harness & 10.00 & 5.00 & 15.00 & 10.00 & 5.00 & 0.00 & 10.00 & 0.00 \\
\bottomrule
\end{tabular}
}
\end{table*}

We further compare recovery performance across base models, agent harnesses, and scientific task families. Table~\ref{tab:appendix-core-domain-recovery} reports domain-level Core-set recovery; Appendix~\ref{app:complete-results} provides Full-set and numerical results, Appendix~\ref{app:statistical-comparisons} provides statistical comparisons, and Table~\ref{tab:appendix-family-recovery} reports family-level results.

With Codex fixed, GPT-5.6-sol achieves substantially higher phenomenal-law and mechanism recovery than the two flash models (Table~\ref{tab:main-recovery}), while no significant difference is detected between GLM-5.3-flash and DeepSeek-v4-flash-0731. GPT-5.6-sol also achieves higher accuracy under Gold-P Direct-Ask (Table~\ref{tab:oracle-controls}), which may reflect a stronger ability to infer latent mechanistic structure from observable consequences and scientific context.

In contrast, holding the base model fixed, we do not detect significant performance differences among the evaluated agent harnesses. Similarly, no significant domain-level differences are detected among physics, chemistry, biology, and materials science. However, individual task families exhibit substantially different phenomenal-law and mechanism recovery profiles. For example, \textit{Inclined Rolling} has relatively high phenomenal-law recovery but low mechanism recovery, whereas \textit{Competitive Inhibition Pulse Reaction} shows the opposite pattern. Overall, the observed variation is more pronounced across base models and task families than across harnesses or scientific domains, indicating that mechanism-recovery difficulty is highly heterogeneous across specific task families and is not well characterized by broad domain labels alone.

\section{Conclusion}
\label{sec:conclusion}

We introduced \textsc{MechBench} to evaluate mechanism discovery beyond phenomenal-law recovery. MechBench combines controlled mechanism mutations, mechanism probes, and screening for mechanistic indistinguishability to enable objective evaluation of internal scientific structure beyond observational agreement. Across representative scientific agents, we observe a substantial gap between phenomenal-law and mechanism recovery, with the gap widening as mechanisms depart further from their canonical forms. Even when the correct phenomenal law is provided, mechanism recovery remains difficult, particularly on increasingly mutated mechanisms. Together, these results reveal a substantial generalization gap in mechanistic reasoning and establish mechanism discovery as a distinct challenge beyond recovering observable scientific laws.

A primary limitation of MechBench is that it currently focuses on mechanisms whose observable phenomenal laws can be obtained through tractable symbolic elimination. Many scientific systems do not admit such a closed-form reduction: deriving observable behavior from the underlying mechanism may require solving nonlinear or transcendental equations, integrating differential equations, or performing numerical simulation. As a result, the current benchmark covers only a subset of mechanistic discovery problems for which the mapping $\mathcal M \rightarrow \mathcal P$ can be expressed analytically. A natural extension is to bypass explicit construction of $\mathcal P$ and generate observations directly from the mechanism through simulation, $\mathcal M \rightarrow D$. Such a setting would substantially broaden the scope of mechanism discovery, but would also require new methods for validating submitted mechanisms and evaluating mechanism probes under numerical approximation, as well as revisiting mechanistic indistinguishability when equivalence can no longer be established symbolically.

\PageLimitAuditEnding

\ifdefined\ArxivVersion
  \subsection*{AI use statement}

In this work, we used generative AI tools to assist with research ideation and hypothesis refinement, the development and critique of benchmark methodology and experimental design, the construction and refinement of synthetic benchmark instances, mechanism variants, and mechanism probes, and the implementation and debugging of research code. We have not used generative AI for formulating or proving mathematical theorems or for translation; tasks involving externally collected datasets or qualitative data analysis were not applicable to this work. Additionally, we used generative AI tools for literature search and summarization, brainstorming, improving presentation and readability, and suggesting paper structure and terminology. We have reviewed all AI-assisted work. Specifically, we checked LLM-generated research ideas for potential plagiarism through a manual literature survey, and verified and tested LLM-generated code for correctness. We take responsibility for the final content of this work, including text, claims, or artifacts produced with the aid of generative AI.

\subsection*{Ethics statement}

This work evaluates scientific reasoning in AI systems using controlled benchmark tasks and synthetically generated observations derived from scientific mechanism models. It does not involve human subjects, personal or private data, or decisions affecting individuals. We foresee no ethical concerns.

\subsection*{Reproducibility statement}

\ifdefined\ArxivVersion
    We provide detailed documentation of benchmark construction and experimental evaluation throughout the paper and appendices. Our code, task instances, and data are publicly available at \url{https://github.com/tsinghua-fib-lab/MechBench}.
\else
    We provide detailed documentation of benchmark construction and experimental evaluation throughout the paper and appendices. Our code, task instances, and data are publicly available at \url{https://anonymous.4open.science/r/mdbench-F567/README.md}.
\fi

\else
  
\fi
\bibliography{reference}

\appendix
\section{Task Construction Details}
\label{app:task-construction-details}

\subsection{Task Families and Reference Mechanisms}
\label{app:mechanism-families}

MechBench contains 16 task families, with four families drawn from each of physics, chemistry, biology, and materials science.  Each family is organized around a standard scientific construction.  To ensure that every internal quantity and observable has an unambiguous value, we combine canonical relations with explicit regime assumptions and, where needed, synthetic but scientifically interpretable closures.  Table~\ref{tab:task-family-overview} summarizes the scientific blueprint of each Original instance.

\begin{table*}[t]
    \centering
    \caption{The 16 task families and the scientific constructions used for their Original reference mechanisms.}
    \label{tab:task-family-overview}
    \resizeTabular{\begin{tabular}{@{}p{3.0cm}p{6.0cm}p{16.0cm}@{}}
\toprule
Domain & Task family & Description \\
\midrule
Physics
& Circular Orbit
& Models the orbital period of a test body using Newtonian inverse-square attraction, circular radial balance, and the speed--period relation. \\
& Hall Response
& Models the voltage measured across offset contacts using a classical carrier-conductivity tensor and an open transverse circuit. \\
& Inclined Rolling
& Models the downslope acceleration of a rigid body from coupled translational and rotational balances under no-slip rolling. \\
& Terminal Settling
& Models the terminal speed of a sphere from gravity, buoyancy, and linear Stokes drag at low Reynolds number. \\
\midrule
Chemistry
& Acid--Base Buffer Relaxation
& Models transient pH response through acid--base buffer capacities and coupled proton-exchange population balances. \\
& Competitive Inhibition Pulse Reaction
& Models product response through mass-action balances among free enzyme, substrate-bound enzyme, and inhibitor-bound enzyme. \\
& Lindemann Pulse Reaction
& Models product response with a collisionally activated Lindemann scheme containing activation, deactivation, and product formation. \\
& Surface Pair Catalysis
& Models catalytic product response through adsorption, desorption, paired surface-complex formation, and surface reaction. \\
\midrule
Biology
& Allosteric Regulation
& Models catalytic activity using a two-conformation statistical occupancy model with conformation-dependent turnover. \\
& Gene Repression
& Models transcription initiation through mutually exclusive promoter occupancies and productive polymerase binding. \\
& Membrane Channel Flux
& Models membrane current density by coupling channel-state occupancy to constant-field electrodiffusive ion flux. \\
& Population Resource Balance
& Models stationary population density using chemostat resource balances and saturating resource-limited growth. \\
\midrule
Materials
& Composite Heat Transport
& Models heat flux by one-dimensional Fourier transport through two material layers in series. \\
& Grain-Boundary Electrical Response
& Models effective conductivity through series grain-interior and grain-boundary complex admittances in a brick-layer geometry. \\
& Thermoelastic Heating Response
& Models stress from preparation-dependent thermal relaxation, thermal eigenstrain, and elastic bar--frame compatibility. \\
& Vacancy Diffusion
& Models diffusivity from equilibrium vacancy formation and thermally activated jumps in a three-dimensional random walk. \\
\bottomrule
\end{tabular}
}
\end{table*}

\paragraph{Physics.} The four Original mechanisms instantiate familiar physical models.  \textit{Circular Orbit} \citep{Newton1687} uses Newtonian inverse-square gravity and circular-motion balance to model the orbital period of a test body.  \textit{Hall Response} \citep{Hall1879} uses the classical carrier-transport tensor under an open transverse circuit, together with an offset contact geometry, to model a measured Hall voltage.  \textit{Inclined Rolling} \citep{Goldstein2002} uses the coupled translational and rotational dynamics of a rigid body subject to a no-slip constraint.  \textit{Terminal Settling} \citep{Stokes1851} uses the steady gravity--buoyancy balance of a sphere in the low-Reynolds-number Stokes regime.

\paragraph{Chemistry.} The four Original mechanisms model transient chemical responses using state-population balances.  \textit{Acid--Base Buffer Relaxation} \citep{Eigen1964} uses proton exchange between the solution and buffer sites to model the pH response to a pulse.  \textit{Competitive Inhibition Pulse Reaction} \citep{Segel1975} uses transitions among free, substrate-bound, and inhibitor-bound enzyme states to model product formation under competition.  \textit{Lindemann Pulse Reaction} \citep{Lindemann1922} uses collisional activation, deactivation, and product formation to model the response of a unimolecular reaction.  \textit{Surface Pair Catalysis} \citep{Langmuir1922} uses adsorption, desorption, paired surface-complex formation, and surface reaction to model catalytic product formation.

\paragraph{Biology.} The four Original mechanisms represent distinct forms of biological regulation and transport.  \textit{Allosteric Regulation} \citep{Monod1965} uses ligand occupancies across two protein conformations, with conformation-dependent catalytic turnover, to model activity.  \textit{Gene Repression} \citep{Bintu2005} uses mutually exclusive promoter occupancies for repressor and polymerase to model transcription initiation.  \textit{Membrane Channel Flux} \citep{Goldman1943} couples closed and open channel states to constant-field electrodiffusion to model ionic current density.  \textit{Population Resource Balance} \citep{Monod1949,NovickSzilard1950} uses steady chemostat balances and saturating resource-limited growth to model population density.

\paragraph{Materials science.} The four Original mechanisms describe transport and coupled thermomechanical responses in materials.  \textit{Composite Heat Transport} \citep{Fourier1822} uses one-dimensional Fourier conduction through two layers in series to model heat flux.  \textit{Grain-Boundary Electrical Response} \citep{Bauerle1969,Macdonald1987} uses grain-interior and grain-boundary admittances in a brick-layer geometry to model effective conductivity.  \textit{Thermoelastic Heating Response} \citep{BoleyWeiner1960} combines preparation-dependent thermal relaxation, thermal eigenstrain, and bar--frame compatibility to model stress.  \textit{Vacancy Diffusion} \citep{Mehrer2007} combines equilibrium vacancy formation with thermally activated atomic jumps in a three-dimensional random walk to model diffusivity.

\subsection{Full-set and Core-set Membership}
\label{app:set-membership}

For each of the 16 families, the Full-set contains the Original mechanism and all 31 nonempty combinations of five family-specific base mutations, for 32 instances per family and 512 instances in total.  The Core-set is a fixed subset containing two families per domain: Circular Orbit and Hall Response (physics); Lindemann Pulse Reaction and Surface Pair Catalysis (chemistry); Allosteric Regulation and Membrane Channel Flux (biology); and Thermoelastic Heating Response and Vacancy Diffusion (materials science).  Each selected family contributes ten existing Full-set instances: the Original, two single-mutation, two double-mutation, two triple-mutation, two four-mutation, and one five-mutation instance.  Thus the Core-set contains 80 instances and is stratified by mutation count rather than generated as a separate collection.

\subsection{Mechanism Mutations and Their Composition}
\label{app:mechanism-mutations}

A base mutation changes one or more relations within the reference mechanism.  The phenomenal law for the resulting variant is then derived from the modified mechanism rather than edited directly.  We chose each mutation to admit a concrete scientific reading and to preserve the dimensional consistency and operating regime of its family.  The same mathematical form can have different meanings in different families, and conversely a single scientific change may require several equations.  For example, introducing a reversible chemical intermediate requires a new state, forward and reverse fluxes, and corresponding changes to multiple population balances.  We group the changes descriptively in Table~\ref{tab:mutation-types}; these categories provide directional guidance only, do not cover every possible mechanism mutation, and may not apply to every task family.

\begin{table*}[t]
    \centering
    \begin{tabular}{@{}p{2.6cm}p{4.3cm}p{6.0cm}@{}}
    \toprule
    Mutation type & Scientific interpretation & Representative changes in the benchmark \\
    \midrule
    Constitutive or component response
    & Change how a component property depends on its local state or an observable control.
    & Mobile carrier trapping in Hall Response; temperature-difference-dependent conductivity in Composite Heat Transport; pressure-dependent migration barrier in Vacancy Diffusion. \\
    Internal state or storage
    & Add a scientifically meaningful population, compartment, reservoir, or relaxation mode.
    & Product-bound enzyme and surface-product intermediates; inactive channel state; slow thermal and anelastic reservoirs. \\
    Additional transport or reaction channel
    & Open a parallel pathway that contributes a new flux while retaining the original pathway.
    & Finite-range orbital attraction, conductive heat bypass, fast grain-boundary path, and alternate vacancy-migration channel. \\
    Source, sink, or balance modification
    & Add a force, production, loss, or exchange term to a governing balance.
    & Resisting torque in rolling, external lift in settling, direct Lindemann product formation, and feed-dependent mortality. \\
    Geometric or kinematic constraint
    & Change how geometric quantities or motions are related without directly editing the final observable law.
    & Non-Euclidean circumferential radius, controlled rolling slip, conducting-area ratio, and pressure-dependent jump distance. \\
    \bottomrule
\end{tabular}

    \caption{Scientific interpretations of the mechanism mutations.}
    \label{tab:mutation-types}
\end{table*}

\paragraph{Family-specific base mutations.} Every family has five labeled mutations $\Delta_1,\ldots,\Delta_5$.  Tables~\ref{tab:physics-mutations}, \ref{tab:chemistry-mutations}, \ref{tab:biology-mutations}, and \ref{tab:materials-mutations} catalog their scientific content for physics, chemistry, biology, and materials science, respectively.

\begin{table*}[t]
    \centering
    \caption{The five base mutations for each physics task family.}
    \label{tab:physics-mutations}
    \resizeTabular{\begin{tabular}{@{}p{7cm}p{18cm}@{}}
\toprule
Family & Mutation \\
\midrule
\multirow{5}{7cm}{Circular Orbit}
& $\Delta_1$: Make relative inertia depend on radius. \\
& $\Delta_2$: Add a finite-range central attraction. \\
& $\Delta_3$: Modify circumferential geometry together with its radial derivative. \\
& $\Delta_4$: Add an external central trapping field. \\
& $\Delta_5$: Store tangential momentum in a coupled internal reservoir. \\
\midrule
\multirow{5}{7cm}{Hall Response}
& $\Delta_1$: Trap a density-dependent fraction of electrons. \\
& $\Delta_2$: Change the electron magnetic-deflection response. \\
& $\Delta_3$: Add a minority-hole transport channel. \\
& $\Delta_4$: Change the transverse electric-force response. \\
& $\Delta_5$: Add a longitudinal bypass carrying no transverse Hall current. \\
\midrule
\multirow{5}{7cm}{Inclined Rolling}
& $\Delta_1$: Make the body's rotational inertia angle-dependent. \\
& $\Delta_2$: Replace no-slip rolling with a controlled slip ratio. \\
& $\Delta_3$: Couple an internal rotor to body rotation. \\
& $\Delta_4$: Add an angle-dependent resisting torque. \\
& $\Delta_5$: Apply an external upslope tangential force. \\
\midrule
\multirow{5}{7cm}{Terminal Settling}
& $\Delta_1$: Add a saturating surface-friction channel. \\
& $\Delta_2$: Reduce displaced fluid volume through accessible pores. \\
& $\Delta_3$: Change the particle-local fluid viscosity. \\
& $\Delta_4$: Apply a radius-dependent external upward force. \\
& $\Delta_5$: Add a parallel dissipative internal-flow channel. \\
\bottomrule
\end{tabular}
}
\end{table*}

\begin{table*}[t]
    \centering
    \caption{The five base mutations for each chemistry task family.}
    \label{tab:chemistry-mutations}
    \resizeTabular{\begin{tabular}{@{}p{7cm}p{18cm}@{}}
\toprule
Family & Mutation \\
\midrule
\multirow{5}{7cm}{Acid--Base Buffer Relaxation}
& $\Delta_1$: Split buffer sites between two acid families. \\
& $\Delta_2$: Introduce a slowly exchanging interior buffer region. \\
& $\Delta_3$: Add a diffusively coupled solvent pocket. \\
& $\Delta_4$: Add proton-catalyzed exchange without changing equilibrium capacity. \\
& $\Delta_5$: Insert a proton-storing relay between solution and primary buffer. \\
\midrule
\multirow{5}{7cm}{Competitive Inhibition Pulse Reaction}
& $\Delta_1$: Add substrate-assisted catalytic release. \\
& $\Delta_2$: Add inhibitor-assisted association to the inhibited state. \\
& $\Delta_3$: Insert a product-bound enzyme intermediate. \\
& $\Delta_4$: Allow slow reversible locking of inhibitor-bound enzyme. \\
& $\Delta_5$: Allow a mixed substrate--inhibitor enzyme complex. \\
\midrule
\multirow{5}{7cm}{Lindemann Pulse Reaction}
& $\Delta_1$: Add a two-collision irreversible deactivation channel. \\
& $\Delta_2$: Insert a second reactive activated intermediate. \\
& $\Delta_3$: Allow reversible nonreactive cage storage before activation. \\
& $\Delta_4$: Add a reversible two-collider activation--deactivation channel. \\
& $\Delta_5$: Open a direct product-forming channel from unactivated reactant. \\
\midrule
\multirow{5}{7cm}{Surface Pair Catalysis}
& $\Delta_1$: Insert a product-bearing surface intermediate. \\
& $\Delta_2$: Allow reversible immobilization of singly adsorbed A. \\
& $\Delta_3$: Add reversible collision-assisted A exchange beside adsorbed B. \\
& $\Delta_4$: Allow gas-A-assisted reversible exchange at the B site. \\
& $\Delta_5$: Open a gas--surface reaction channel from singly adsorbed B. \\
\bottomrule
\end{tabular}
}
\end{table*}

\begin{table*}[t]
    \centering
    \caption{The five base mutations for each biology task family.}
    \label{tab:biology-mutations}
    \resizeTabular{\begin{tabular}{@{}p{7cm}p{18cm}@{}}
\toprule
Family & Mutation \\
\midrule
\multirow{5}{7cm}{Allosteric Regulation}
& $\Delta_1$: Add an inactive conformation binding three regulators. \\
& $\Delta_2$: Couple inactive-state substrate affinity to substrate exposure. \\
& $\Delta_3$: Add another productive active-state complex. \\
& $\Delta_4$: Partition regulator into a local compartment. \\
& $\Delta_5$: Change inactive-conformation catalytic turnover. \\
\midrule
\multirow{5}{7cm}{Gene Repression}
& $\Delta_1$: Add a looped state containing two repressors. \\
& $\Delta_2$: Allow a weakly productive repressor--polymerase complex. \\
& $\Delta_3$: Change repressor affinity with polymerase exposure. \\
& $\Delta_4$: Partition polymerase into a promoter-local compartment. \\
& $\Delta_5$: Add a repressor-dependent paused-polymerase state. \\
\midrule
\multirow{5}{7cm}{Membrane Channel Flux}
& $\Delta_1$: Make channel opening voltage-dependent. \\
& $\Delta_2$: Add an extracellular-concentration-dependent inactivated state. \\
& $\Delta_3$: Partition the intracellular ion compartment. \\
& $\Delta_4$: Make channel permeability voltage-dependent. \\
& $\Delta_5$: Introduce a second monovalent species with proportionally clamped reservoirs. \\
\midrule
\multirow{5}{7cm}{Population Resource Balance}
& $\Delta_1$: Add feed-dependent mortality. \\
& $\Delta_2$: Partition organisms between active and protected compartments. \\
& $\Delta_3$: Add resource-consuming maintenance. \\
& $\Delta_4$: Reduce maximum birth frequency at high feed concentration. \\
& $\Delta_5$: Partition feed resource into usable and inaccessible pools. \\
\bottomrule
\end{tabular}
}
\end{table*}

\begin{table*}[t]
    \centering
    \caption{The five base mutations for each materials-science task family.}
    \label{tab:materials-mutations}
    \resizeTabular{\begin{tabular}{@{}p{7cm}p{18cm}@{}}
\toprule
Family & Mutation \\
\midrule
\multirow{5}{7cm}{Composite Heat Transport}
& $\Delta_1$: Make the first component's conductivity depend on the imposed temperature difference. \\
& $\Delta_2$: Add a composition-dependent thermal interface resistance. \\
& $\Delta_3$: Introduce a parallel conductive bypass. \\
& $\Delta_4$: Change the second component's conducting area. \\
& $\Delta_5$: Apply controlled volumetric heating within the second component. \\
\midrule
\multirow{5}{7cm}{Grain-Boundary Electrical Response}
& $\Delta_1$: Add a thin space-charge region beside each boundary. \\
& $\Delta_2$: Add delayed boundary polarization. \\
& $\Delta_3$: Allow a parallel fast boundary path. \\
& $\Delta_4$: Add delayed polarization inside grains. \\
& $\Delta_5$: Include an additional boundary-local interface impedance. \\
\midrule
\multirow{5}{7cm}{Thermoelastic Heating Response}
& $\Delta_1$: Add a second thermal equilibration mode. \\
& $\Delta_2$: Add a recoverable anelastic strain reservoir. \\
& $\Delta_3$: Include finite elastic compliance of an adhesive layer. \\
& $\Delta_4$: Add a slow thermal-expansion reservoir. \\
& $\Delta_5$: Add a second anelastic strain-relaxation time. \\
\midrule
\multirow{5}{7cm}{Vacancy Diffusion}
& $\Delta_1$: Trap a temperature-dependent fraction of vacancies. \\
& $\Delta_2$: Add a quadratic pressure contribution to the migration barrier. \\
& $\Delta_3$: Contract the effective atomic jump distance. \\
& $\Delta_4$: Introduce correlated return jumps. \\
& $\Delta_5$: Open a parallel activated migration channel. \\
\bottomrule
\end{tabular}
}
\end{table*}

\clearpage

\paragraph{Composition rule.} For a family with mutation index set $S\subseteq\{1,\ldots,5\}$, we construct one mechanism $\mathcal M^{(S)}$.  If mutations act on disjoint relations, the modified relations are simply included together.  If they share a state or balance equation, we construct the joint equation from the union of their fluxes or components, so that every contribution occurs once and conservation is maintained.  Shared constants and unchanged relations are likewise merged, not duplicated.  This construction makes the compositions order-independent: the name ``Variant 1-3-5,'' for example, denotes the set $S=\{1,3,5\}$.  The five base mutations in every family were selected and, where necessary, adjusted so that all subsets are compatible; hence the Full-set retains all $2^5=32$ choices of $S$, including $S=\emptyset$ as Original.

We measure variant complexity by $|S|$, the number of selected base mutations.  A single base mutation may modify several related equations, but it still contributes one to $|S|$.  We retain a composed mechanism only when its equations admit a unique explicit algebraic solution for every internal quantity and the target as functions of the input and auxiliary variables.  After composing $\mathcal M^{(S)}$, we eliminate its internal variables to obtain $\mathcal P^{(S)}$ and derive the corresponding probe answers.  All variants are therefore constructed at the mechanism level before elimination.

\paragraph{Composition example: Vacancy Diffusion.} This family provides a compact example.  Omitting unchanged zero-valued channels, the Original mechanism contains
\begin{align}
c_v &= \exp\!\left[-\frac{E_f+p\Omega}{k_BT}\right], &
c_m &= c_v, \\
E_b &= E_m, &
r &= \nu\exp\!\left[-\frac{E_b}{k_BT}\right], &
D &= \frac{a^2c_mr}{6}.
\end{align}
Mutation 1 replaces only the mobile-vacancy relation with
\begin{equation}
c_m=\frac{c_v}{1+3\exp[-E_t/(k_BT)]},
\label{eq:vacancy-mutation-one}
\end{equation}
representing temperature-dependent trapping.  Mutation 2 instead replaces the migration-barrier relation with
\begin{equation}
E_b=E_m+\frac{\Omega p^2}{p_{\mathrm{ref}}},
\label{eq:vacancy-mutation-two}
\end{equation}
representing a nonlinear pressure contribution.  Applying both changes gives the joint mechanism containing Eqs.~\ref{eq:vacancy-mutation-one} and \ref{eq:vacancy-mutation-two}, and elimination yields
\begin{equation}
\mathcal P^{(1,2)}:\quad
D=\frac{a^2\nu}{6}
\frac{\exp[-(E_f+p\Omega)/(k_BT)]
      \exp[-(E_m+\Omega p^2/p_{\mathrm{ref}})/(k_BT)]}
     {1+3\exp[-E_t/(k_BT)]}.
\label{eq:vacancy-composed-phenomenon}
\end{equation}
Thus the combined phenomenal law is a consequence of two changes made at different mechanistic locations.  We probe the mobile vacancy fraction $c_m$ and the migration barrier $E_b$ separately, so matching their combined effect on $D$ is not sufficient to recover the mechanism.

\subsection{Mechanism Probe Construction}
\label{app:probe-construction}

\paragraph{Selection principles.} For each instance, we select scientifically interpretable internal quantities whose values follow from the reference mechanism but are not fixed by the phenomenal law and the provided scientific context alone.  A probe is specified by a verbal description of the quantity and a reference expression $z=g_{\mathcal M}(\mathbf x)$.  The expression is obtained by solving the mechanism relations and eliminating the remaining internal variables, so it can ultimately be evaluated from the observable inputs and known numerical constants.  Although the final expression is a function of observable inputs, deriving it still requires internal constitutive, balance, geometric, or dynamical relations absent from $\mathcal P$.

We use probes with clear scientific meanings.  Eligible probe quantities must contribute to the derivation of the phenomenal law: if removing the relations involving a quantity leaves the same phenomenal law derivable, that quantity is omitted.  A probe quantity must also have the same value across comparably plausible mechanisms consistent with the observable evidence; otherwise, scoring it would reward agreement with the designated reference rather than recovery of a mechanism supported by the available evidence.  Mechanistic-indistinguishability screening in Section~\ref{sec:mechanistic-indistinguishability} enforces the latter requirement.

\paragraph{Example.} In the Vacancy Diffusion composition above, the phenomenal law contains the product of the mobile vacancy fraction and an activated jump rate.  The probe for $c_m$ requires the equilibrium vacancy relation and the trapping correction in Eq.~\ref{eq:vacancy-mutation-one}, whereas the probe for $E_b$ requires the pressure-dependent barrier relation in Eq.~\ref{eq:vacancy-mutation-two}.  Their reference answers are
\begin{equation}
c_m=\frac{\exp[-(E_f+p\Omega)/(k_BT)]}{1+3\exp[-E_t/(k_BT)]}, \qquad E_b=E_m+\frac{\Omega p^2}{p_{\mathrm{ref}}}.
\end{equation}
Both are functions of observable inputs and known constants after the mechanism has been applied, but neither factorization is determined by the single expression for $D$ alone.  Querying them separately therefore distinguishes recovery of the trapping and pressure-barrier relations from recovery of their aggregate observable effect.

\paragraph{Variation across instances.} Probe sets are adapted to the active mechanism relations rather than held fixed across a family.  The Original instance uses a representative internal quantity from its reference mechanism.  A single mutation may change that quantity or replace it with a more diagnostic internal consequence; a composition may require several probes to cover changes at distinct mechanistic locations.  We do not require one probe per mutation because one internal quantity can reflect several coupled changes, while some mutations introduce more than one scientifically relevant consequence.  Across the 512 Full-set instances, $K$ ranges from one to five: 94 instances have one probe, 314 have two, 68 have three, 30 have four, and six have five.

\subsection{Observation Generation and Task Specification}
\label{app:observation-generation}

\paragraph{Reference expressions and valid domains.} Before generating observations, we solve the mechanism equations for every internal variable and the target as explicit functions of the input and auxiliary variables.  For each observable source variable $x_j$, we choose an instance-specific lower limit $l_j$, extrapolation boundary $b_j$, upper limit $u_j$, and either a uniform or log-uniform sampling distribution according to its scientific meaning, characteristic scale, and physically meaningful operating regime.  The resulting ranges also avoid known singularities or invalid regimes.

\paragraph{ID and OOD sampling.} We independently sample all source variables using their specified distributions.  Training observations and the held-out ID test set use $x_j\in[l_j,b_j)$ for every $j$, whereas the OOD test set uses $x_j\in[b_j,u_j)$ for every $j$.  Thus ID testing uses new draws from the same region as training, while OOD testing evaluates joint high-side extrapolation across the source variables.  The default configuration, used throughout the benchmark, is summarized in Table~\ref{tab:observation-generation}.

\begin{table}[t]
    \centering
    \begin{tabular}{lccc}
        \toprule
        Split & Region for each $x_j$ & Samples & Agent access \\
        \midrule
        Training & $[l_j,b_j)$ & 1,000 & Yes \\
        ID test & $[l_j,b_j)$ & 1,000 & No \\
        OOD test & $[b_j,u_j)$ & 1,000 & No \\
        \bottomrule
    \end{tabular}
    \caption{Default observation-generation configuration.}
    \label{tab:observation-generation}
\end{table}

The target and all internal quantities are evaluated from the solved reference mechanism for each sampled input.  Samples producing a non-real or non-finite value for any derived quantity are discarded and resampled.  Generation uses a task-local pseudorandom generator with default seed 0.  No observational noise is added.

\paragraph{Information exposed during discovery.} The agent receives a short scientific task description; the name, description, unit, and role of each observable input and target; the row order of the data array; and the training observations.  The array has shape $(n_{\mathrm{variables}},1000)$, with the target in the first row followed by the input and auxiliary variables.  The task description is identical across all variants of a family, and neither the instance identifier nor the active mutations are disclosed.  Sampling limits and distributions, internal variables, the reference $\mathcal M$ and $\mathcal P$, mechanism-probe definitions and answers, and the two held-out test sets remain inaccessible during discovery.  For example, the discovery view for the Vacancy Diffusion composition in Eqs.~\ref{eq:vacancy-mutation-one}--\ref{eq:vacancy-composed-phenomenon} contains the metadata shown below together with a four-row training array in the listed variable order.

\begin{center}
    \begin{minipage}{0.98\linewidth}
    \footnotesize\ttfamily\raggedright
    \{\\
    \hspace*{1em}"task\_description": "Determine atomic diffusivity from temperature, hydrostatic pressure, and lattice spacing.",\\
    \hspace*{1em}"variables": [\\
    \hspace*{2em}\{"name": "diffusivity", "description": "Atomic tracer diffusivity.",\\
    \hspace*{3em}"unit": "m\textasciicircum2 s\textasciicircum-1", "role": "target"\},\\
    \hspace*{2em}\{"name": "temp", "description": "Temperature.",\\
    \hspace*{3em}"unit": "K", "role": "input"\},\\
    \hspace*{2em}\{"name": "pressure", "description": "Hydrostatic pressure.",\\
    \hspace*{3em}"unit": "Pa", "role": "input"\},\\
    \hspace*{2em}\{"name": "spacing", "description": "Lattice spacing.",\\
    \hspace*{3em}"unit": "m", "role": "input"\}\\
    \hspace*{1em}],\\
    \hspace*{1em}"data\_columns": ["diffusivity", "temp", "pressure", "spacing"],\\
    \hspace*{1em}"data\_layout": "variables\_by\_samples"\\
    \}
    \end{minipage}
\end{center}

The agent must infer the observable law and its internal explanation from the shared scientific context and observations; variant-specific metadata is not exposed.

\paragraph{Reference feedback script.} During the search process, the agent may repeatedly run a feedback script on a candidate set of equations together with the public metadata and training observations.  The script is provided as a reference implementation that helps the agent understand the required mechanism-submission format, how the submitted equations are algebraically eliminated to obtain the observable target, and how the resulting expression is evaluated on the training set.  For a valid candidate, it returns the explicit solution implied by the submitted equations and numerical fit metrics computed on the training observations; malformed, inconsistent, or unsolved candidates instead return an error.  It reveals no information beyond the already available task description and training data, and does not access or report the reference mechanism, the reference phenomenal law, mechanism probes, or held-out test sets.

\paragraph{Post-search probe queries.} After the search process ends, the submitted mechanism is locked, and each mechanism probe is presented in an independent query.  The query provides the scientific description of the requested probe quantity, but not the reference probe answer or any held-out observations.  Separating this later query from the discovery interface prevents the probe itself from revealing the intended internal structure while the agent is constructing its mechanism.

\section{Experimental Details}
\label{app:experimental-details}

\subsection{Evaluated Configurations}

Table~\ref{tab:eval-configs} summarizes the model--agent configurations and control conditions
used in our experiments.
All seven standard-discovery configurations are evaluated on the Core-set.
For the Full-set, we evaluate Codex with GPT-5.6-sol and GLM-5.3-flash.
PySR, PySR + Direct-Ask, and the Gold-$P$ Direct-Ask controls are evaluated on both sets.

\begin{table}[t]
\centering
\small
\begin{tabular}{lllcc}
\toprule
\textbf{Setting} &
\textbf{Base Model} &
\textbf{Agent / Pipeline} &
\textbf{Core} &
\textbf{Full} \\
\midrule
Standard discovery & GPT-5.6-sol               & Codex            & \checkmark & \checkmark \\
Standard discovery & GLM-5.3-flash             & Codex            & \checkmark & \checkmark \\
Standard discovery & DeepSeek-v4-flash-0731    & Codex            & \checkmark &            \\
Standard discovery & GLM-5.3-flash             & Claude Code      & \checkmark &            \\
Standard discovery & DeepSeek-v4-flash-0731    & Claude Code      & \checkmark &            \\
Standard discovery & DeepSeek-v4-flash-0731    & DeepSeek Harness & \checkmark &            \\
Standard discovery & DeepSeek-v4-pro-0813      & DeepSeek Harness & \checkmark &            \\
\midrule
Equation baseline  & --                         & PySR             & \checkmark & \checkmark \\
Pipeline baseline  & DeepSeek-v4-flash-0731    & PySR + Direct-Ask& \checkmark & \checkmark \\
\midrule
Gold-$P$            & GPT-5.6-sol               & Direct-Ask       & \checkmark & \checkmark \\
Gold-$P$            & GLM-5.3-flash             & Direct-Ask       & \checkmark & \checkmark \\
Gold-$P$            & DeepSeek-v4-flash-0731    & Direct-Ask       & \checkmark & \checkmark \\
\bottomrule
\end{tabular}
\caption{Model--agent configurations, baselines, and controls used in the experiments.}
\label{tab:eval-configs}
\end{table}

For standard discovery, Core-set results for Codex with GLM-5.3-flash and
DeepSeek-v4-flash-0731 are averaged over three independent runs.
All other reported standard-discovery configurations use a single run.
For each repeated configuration, metrics are first computed separately using the fixed task
denominator and are then averaged across runs rather than pooling predictions across runs.

\subsection{Standard Discovery and Baselines}

\paragraph{Standard discovery.}
All agents interact with the same benchmark interface described in Appendix~\ref{app:observation-generation}.
For each task, the agent receives the scientific task description, observable-variable metadata,
and training observations $D$.
During search, it may evaluate candidate mechanisms using the reference feedback script, which
solves the submitted equations when possible and reports objective fit information computed only
from the training observations.
The script does not access the reference mechanism, the reference phenomenal law, mechanism
probes, or held-out ID/OOD observations.
After search terminates, the final mechanism submission is locked before any mechanism probe is
revealed.

\paragraph{PySR.}
We use PySR as a non-LLM baseline for phenomenal-law discovery.
PySR receives the same observable training data and attempts to recover an equation relating the
observable inputs to the target.
Its recovered equation is evaluated against the reference phenomenal law using the same
phenomenal-law evaluation procedure as agent submissions.

\paragraph{PySR + Direct-Ask.}
To test whether an equation recovered by a conventional symbolic-regression method can support
subsequent mechanism reasoning, we additionally evaluate a two-stage PySR + Direct-Ask pipeline.
PySR first discovers a phenomenal equation from the training observations.
DeepSeek-v4-flash-0731 is then provided with this recovered equation and independently answers
the mechanism probes.

\paragraph{Gold-$P$ Direct-Ask.}
Gold-$P$ Direct-Ask removes phenomenal-law discovery entirely.
For each task, the model is provided with the reference phenomenal law $P$ together with the
numerical constants required to interpret the requested quantity.
Each mechanism probe is asked in an independent, tool-free query.
The model is not provided with the reference mechanism or the reference probe expression.
Success therefore depends on reconstructing enough internal scientific structure to derive the
requested quantity from the observable law and context.

\subsection{Evaluation Metrics}

Let $P_i$ and $\widehat{P}_i$ denote the reference and predicted phenomenal laws for task $i$,
and let $g_{ik}$ and $\widehat{g}_{ik}$ denote the reference and predicted expressions for its
$k$-th mechanism probe.
Writing $\equiv$ for symbolic equivalence and $K_i$ for the number of probes associated with
task $i$, we define
\begin{equation}
\mathrm{SA}(P)
=
\frac{1}{N}
\sum_{i=1}^{N}
\mathbf{1}\!\left[\widehat{P}_i \equiv P_i\right],
\end{equation}
and
\begin{equation}
\mathrm{SA}(M)
=
\frac{1}{N}
\sum_{i=1}^{N}
\prod_{k=1}^{K_i}
\mathbf{1}\!\left[\widehat{g}_{ik} \equiv g_{ik}\right].
\end{equation}
Thus, $\mathrm{SA}(M)$ is a strict task-level mechanism-recovery metric: a task is counted as
correct only when all of its mechanism probes are symbolically correct.
It is unconditional with respect to phenomenal-law recovery, so a mechanism score is evaluated
regardless of whether $\widehat{P}_i$ is correct.

For the Core-set, we additionally report numerical agreement on the held-out ID and OOD splits.
For $S\in\{\mathrm{ID},\mathrm{OOD}\}$, let $y^{S}_{ij}$ and $\widehat{y}^{S}_{ij}$ denote the
reference and predicted target values for held-out sample $j$ of task $i$.
We define
\begin{equation}
\mathrm{Acc}^{S}_{0.1}
=
\frac{1}{N}
\sum_{i=1}^{N}
\mathbf{1}
\left[
\max_{j\in S_i}
\left|
\frac{\widehat{y}^{S}_{ij}-y^{S}_{ij}}
     {y^{S}_{ij}}
\right|
\leq 0.1
\right].
\end{equation}
This is a task-level accuracy rather than the fraction of individual samples within tolerance:
a task is counted as correct only if its maximum relative error over the entire held-out split does
not exceed $10\%$.

\subsection{Mutation Strata and Reported Subsets}

We report results both over the complete benchmark subsets and after stratifying instances by
mechanism familiarity.
The \emph{Original} instance of each task family has mutation count zero, while constructed
variants contain between one and five applied base mutations.
The Full-set contains one Original and all 31 nonempty mutation combinations for each of its
16 task families.
The Core-set is a fixed subset stratified by mutation count, as detailed in Appendix~\ref{app:set-membership}.

For variant-only analyses, we exclude the Original instance from every included task family,
yielding 496 Full-set variants and 72 Core-set variants.
Results stratified by mutation count and the complete variant-only results are reported in Appendix~\ref{app:variant-results}.

\subsection{Implementation and Runtime Settings}

\paragraph{Search and wall-clock budgets.} Standard discovery uses a 900-second wall-clock limit per task for Codex, Claude Code, DeepSeek Harness, and PySR. Codex, Claude Code, and DeepSeek Harness have no benchmark-imposed maximum number of agent turns and no cap on calls to the observable-fit feedback script; search is limited by wall-clock time instead. In particular, the reported DeepSeek Harness runs set \texttt{dsh-feedback-evaluations=0}, selecting the time-based prompt rather than the experimental fixed-evaluation prompt. No hard monetary budget is enabled for any reported agent configuration. Mechanism-probe turns following standard discovery have a 120-second limit each and are executed independently from the same frozen discovery result.

\paragraph{Reasoning settings.} Codex uses reasoning effort \texttt{xhigh} with GPT-5.6-sol, \texttt{high} with GLM-5.3-flash, and \texttt{low} with DeepSeek-v4-flash-0731. Claude Code uses \texttt{high} with GLM-5.3-flash and \texttt{low} with DeepSeek-v4-flash-0731. Both DeepSeek Harness configurations use \texttt{low}. These settings control the provider's reasoning mode rather than the number of benchmark feedback evaluations.

\paragraph{Codex invocation.} Each discovery task launches \texttt{codex exec} with \texttt{--json}, \texttt{--skip-git-repo-check}, \texttt{--color never}, the selected \texttt{--model}, and \texttt{model\_reasoning\_effort}; the prompt is supplied on standard input and the final message is written with \texttt{-o}. The benchmark additionally sets \texttt{approval\_policy="never"}, disables web search, apps, plugins, hooks, multi-agent operation, memories, and shell snapshots, and installs a per-task permission policy. That policy grants write access only to the temporary task workspace, denies the repository, answers, credentials, and shell startup files, and allows network access only to the loopback feedback server. Probe turns use \texttt{codex exec resume} on an independent copy of the frozen discovery transcript with a read-only sandbox and all shell, code-execution, browser, image, and computer tools disabled.

\paragraph{Claude Code invocation.} Discovery uses \texttt{claude --print --verbose --output-format stream-json --include-partial-messages --input-format text --setting-sources '' --strict-mcp-config --disable-slash-commands --permission-mode dontAsk}, an empty MCP configuration, and \texttt{--tools Bash,Read,Write,Edit,Glob,Grep}, followed by the selected \texttt{--model} and \texttt{--effort}. A generated settings file disables hooks and memory, denies access outside the isolated task workspace, protects \texttt{problem.json} and \texttt{train.npy} from writes, and restricts network access to the feedback host. Each probe uses \texttt{--resume <frozen-session> --fork-session} with \texttt{--tools ''}; consequently probe branches cannot alter either the frozen model or one another. The optional Claude Code \texttt{--max-budget-usd} setting is unset in the reported experiments.

\paragraph{DeepSeek Harness invocation.} Discovery uses \texttt{dsh --profile headless --patch <generated-patch> <task-prompt>}. The generated patch fixes the requested OpenRouter model, disables web search/fetch, subagents, workflows, and related tools, and routes model requests through the benchmark's metered loopback gateway. Probe profiles additionally disable filesystem, shell, job, skill, todo, and goal tools. The two reported models are \texttt{deepseek/deepseek-v4-flash-0731} and \texttt{deepseek/deepseek-v4-pro-0813}; OpenRouter failures do not fall back to a different provider. The live discovery checkpoint is \texttt{working\_model.txt}. The gateway records token and monetary usage but enforces no hard dollar limit.

\paragraph{Prompt comparability.} The three harnesses use prompts with the same scientific objective, observed data, mechanism requirements, access to the observable-fit feedback script, equation-only submission contract, 900-second budget, and isolation rules. Harness-specific wording and checkpoint conventions remain; DeepSeek Harness, for example, refers to \texttt{working\_model.txt} rather than \texttt{submission.txt}. Each CLI also retains its native system instructions and tool descriptions. After discovery, all three standard agents receive the same benchmark-generated mechanism-probe task prompt, and every probe is evaluated in an independent branch from the frozen submitted model.

\paragraph{Sampling, context, and output limits.} The standard Codex and Claude Code launches do not explicitly set temperature, top-$p$, maximum context length, or maximum output tokens; these remain the selected CLI/provider defaults and no uniform values are available in the run receipts. DeepSeek Harness likewise does not override temperature or top-$p$, but caps each model response at 16,000 output tokens.

\paragraph{PySR hyperparameters.} PySR is run for the same 900-second wall-clock budget with \texttt{niterations=1,000,000}, so the timeout rather than the nominal iteration count normally terminates search. We use 31 populations, maximum expression size 30, 64-bit internal precision, multithreaded parallelism, and \texttt{model\_selection="accuracy"}; population size and other unlisted evolutionary settings retain PySR defaults. Binary operators are $+$, $-$, $\times$, division, and $|x|^y$. Unary operators are square, cube, $\sqrt{|x|}$, $\exp(x)$, $\log|x|$, $\sin(x)$, $\cos(x)$, and $|x|$. The exponent operator uses the PySR constraint \texttt{(-1,3)}, and \texttt{exp} uses constraint 8. Nested exponentials and nested sine/cosine compositions are disallowed. Each task's recorded batch seed is passed to PySR as its random state; seeds are task-specific rather than fixed to a common value.

\subsection{Direct-Ask Prompt and Runtime}

\paragraph{Gold-$P$ prompt template.} Gold-$P$ replaces the frozen submission in the common mechanism-probe template with the exact reference phenomenal equation after eliminating internal variables. It also discloses named numerical-constant equations and numerical literals labelled by the mechanism quantity whose defining equation contains them. The user-message template is:

\begin{verbatim}
Frozen submitted phenomenal model:
{reference phenomenal equation}

Known named numerical constants for this task:
{name = numeric value, or (none)}

Additional numerical constant values, labelled by the mechanism
quantity whose defining equation contains them (without disclosing
the equation itself):
{quantity: numeric literals, or (none)}

Using the phenomenal model and known numerical constants above,
derive {probe name} ({probe description}) as a function of:
{variables}.

Return exactly one equation in the form:
{probe name} = <expression>

Use only the given variables, numeric literals, and supported
operators/functions. Do not add variables, equations, or prose;
do not modify the frozen phenomenal model; and do not use tools.
\end{verbatim}

\paragraph{Direct-Ask execution.} Each probe is a fresh one-turn, tool-free query; probe answers do not share conversational state. The GPT-5.6-sol control uses Codex with medium reasoning effort and a 300-second timeout. The GLM-5.3-flash control uses direct OpenRouter chat-completions requests with \texttt{temperature=0}, minimal reasoning effort, an 8,192-token output cap, a 600-second request timeout, and up to four transport attempts. The reported DeepSeek-v4-flash-0731 Gold-$P$ control uses \texttt{temperature=0}, minimal reasoning effort, a 2,048-token output cap, a 180-second request timeout, and up to four transport attempts. The PySR + Direct-Ask pipeline uses the same direct-API system instruction and decoding contract as the latter, also with \texttt{temperature=0}, minimal reasoning effort, a 2,048-token output cap, a 180-second request timeout, and up to four attempts.

\paragraph{Direct-Ask system instruction.} For direct OpenRouter requests, the system message is: ``Answer the user's mechanism probe directly from the supplied frozen equation. Follow its output contract exactly: emit one plain-text equation and no reasoning, prose, Markdown, or additional equations.'' The Codex version appends the same one-equation contract to the user prompt and disables filesystem, network, shell, web, and other tools while retaining its native system scaffolding.

\paragraph{Agent failures.} A run is treated as an agent-side failure when the agent execution itself is available for evaluation but its answer is unusable or scientifically wrong: examples include an empty checkpointed working model, malformed or unparsable equations, a model that cannot jointly solve the required variables, multiple or prose-bearing Direct-Ask answers, and an incorrect numerical or symbolic result. We re-run these failed cases up to five times until they succeed; runs that fail to submit a valid answer after five attempts are recorded as failures.

\section{Experimental Results}

\subsection{Complete Recovery and Numerical-Accuracy Results}
\label{app:complete-results}

Tables~\ref{tab:appendix-core-domain-recovery} and \ref{tab:appendix-full-domain-recovery} give the complete domain-level symbolic-recovery results underlying the aggregate values in Table~\ref{tab:main-recovery}. For the two three-run Codex configurations, the rate is computed separately for each run and then averaged. As in the main text, $\mathrm{SA}(M)$ is the strict task-level measure that requires every mechanism probe for an instance to be symbolically correct.

Codex with GPT-5.6-sol has the highest overall $\mathrm{SA}(P)$ and $\mathrm{SA}(M)$ on both sets and leads mechanism recovery in each Full-set domain. The size of the phenomenal--mechanistic gap nevertheless varies substantially. On the Core-set, for example, this configuration recovers $40.00\%$ of phenomenal laws but only $10.00\%$ of mechanisms in biology, and recovers no complete mechanism in materials despite recovering $20.00\%$ of phenomenal laws.

Table~\ref{tab:appendix-family-recovery} further reports all 16 Full-set task families for the two standard discovery configurations evaluated over the Full-set. A task family that is easier for phenomenal recovery is not necessarily easier for mechanism recovery. For GPT-5.6-sol, \textit{Inclined Rolling} has the highest phenomenal recovery ($50.00\%$) but only $6.25\%$ mechanism recovery, whereas \textit{Competitive Inhibition Pulse Reaction} reaches $31.25\%$ and $25.00\%$, respectively. Conversely, several families have nonzero phenomenal recovery but no completely recovered mechanism.

Finally, Table~\ref{tab:appendix-core-numerical-accuracy} reports the complete Core-set numerical results. Numerical agreement is less stringent than exact symbolic recovery: Codex with GPT-5.6-sol attains $\mathrm{Acc}_{0.1}^{\mathrm{ID}}=55.00\%$ and $\mathrm{Acc}_{0.1}^{\mathrm{OOD}}=43.75\%$, compared with $\mathrm{SA}(P)=35.00\%$. The same separation is visible for the smaller models, whose overall symbolic recovery is at most $10.00\%$ while their OOD numerical accuracy ranges from $16.25\%$ to $25.00\%$. OOD accuracy is lower than ID accuracy for every reported agent configuration.

\begin{table*}[t]
    \centering
    \caption{Complete Full-set symbolic-recovery results by scientific domain.}
    \label{tab:appendix-full-domain-recovery}
    \resizeTabular{\begin{tabular}{llcccccccc}
\toprule
\multirow{2}{*}{\textbf{Base model}} & \multirow{2}{*}{\textbf{Agent / condition}}
& \multicolumn{2}{c}{\textbf{Physics}} & \multicolumn{2}{c}{\textbf{Chemistry}}
& \multicolumn{2}{c}{\textbf{Biology}} & \multicolumn{2}{c}{\textbf{Materials}} \\
\cmidrule(lr){3-4}\cmidrule(lr){5-6}\cmidrule(lr){7-8}\cmidrule(lr){9-10}
& & $\mathcal P$ & $\mathcal M$ & $\mathcal P$ & $\mathcal M$ & $\mathcal P$ & $\mathcal M$ & $\mathcal P$ & $\mathcal M$ \\
\midrule
No LLM & PySR & 0.78 & -- & 0.00 & -- & 0.00 & -- & 0.00 & -- \\
DeepSeek-v4-flash-0731 & PySR + Direct-Ask & 0.78 & 0.78 & 0.00 & 0.00 & 0.00 & 0.00 & 0.00 & 0.00 \\
\midrule
GPT-5.6-sol & Codex & 25.78 & 9.38 & 17.19 & 12.50 & 13.28 & 5.47 & 7.81 & 3.91 \\
GLM-5.3-flash & Codex & 10.16 & 3.12 & 0.78 & 0.00 & 2.34 & 2.34 & 3.91 & 1.56 \\
\bottomrule
\end{tabular}
}
\end{table*}

\begin{table*}[t]
    \centering
    \caption{Full-set symbolic-recovery results by task family for the two standard discovery configurations evaluated on all 512 instances. Each family contains 32 instances.}
    \label{tab:appendix-family-recovery}
    \resizeTabular{\begin{tabular}{llcccc}
\toprule
\multirow{2}{*}{\textbf{Domain}} & \multirow{2}{*}{\textbf{Task family}}
& \multicolumn{2}{c}{\textbf{Codex (GPT-5.6-sol)}}
& \multicolumn{2}{c}{\textbf{Codex (GLM-5.3-flash)}} \\
\cmidrule(lr){3-4}\cmidrule(lr){5-6}
& & $\mathcal P$ & $\mathcal M$ & $\mathcal P$ & $\mathcal M$ \\
\midrule
\multirow{4}{*}{Physics}
& Circular Orbit & 21.88 & 15.63 & 9.38 & 3.12 \\
& Hall Response & 25.00 & 12.50 & 3.12 & 6.25 \\
& Inclined Rolling & 50.00 & 6.25 & 28.12 & 3.12 \\
& Terminal Settling & 6.25 & 3.12 & 0.00 & 0.00 \\
\midrule
\multirow{4}{*}{Chemistry}
& Acid--Base Buffer Relaxation & 3.12 & 3.12 & 0.00 & 0.00 \\
& Competitive Inhibition Pulse Reaction & 31.25 & 25.00 & 3.12 & 0.00 \\
& Lindemann Pulse Reaction & 25.00 & 15.63 & 0.00 & 0.00 \\
& Surface Pair Catalysis & 9.38 & 6.25 & 0.00 & 0.00 \\
\midrule
\multirow{4}{*}{Biology}
& Allosteric Regulation & 18.75 & 3.12 & 0.00 & 0.00 \\
& Gene Repression & 21.88 & 12.50 & 9.38 & 9.38 \\
& Membrane Channel Flux & 9.38 & 3.12 & 0.00 & 0.00 \\
& Population Resource Balance & 3.12 & 3.12 & 0.00 & 0.00 \\
\midrule
\multirow{4}{*}{Materials}
& Composite Heat Transport & 12.50 & 12.50 & 12.50 & 6.25 \\
& Grain-Boundary Electrical Response & 3.12 & 3.12 & 0.00 & 0.00 \\
& Thermoelastic Heating Response & 0.00 & 0.00 & 0.00 & 0.00 \\
& Vacancy Diffusion & 15.63 & 0.00 & 3.12 & 0.00 \\
\bottomrule
\end{tabular}
}
\end{table*}

\begin{table*}[t]
    \centering
    \caption{Core-set numerical accuracy by scientific domain. Each entry is the percentage of tasks whose maximum relative error over the indicated split is at most $0.1$. A dash denotes an unavailable OOD result.}
    \label{tab:appendix-core-numerical-accuracy}
    \resizeTabular{\begin{tabular}{llcccccccc}
\toprule
\multirow{2}{*}{\textbf{Base model}} & \multirow{2}{*}{\textbf{Agent}}
& \multicolumn{2}{c}{\textbf{Physics}} & \multicolumn{2}{c}{\textbf{Chemistry}}
& \multicolumn{2}{c}{\textbf{Biology}} & \multicolumn{2}{c}{\textbf{Materials}} \\
\cmidrule(lr){3-4}\cmidrule(lr){5-6}\cmidrule(lr){7-8}\cmidrule(lr){9-10}
& & ID & OOD & ID & OOD & ID & OOD & ID & OOD \\
\midrule
No LLM & PySR & 35.00 & -- & 0.00 & -- & 20.00 & -- & 0.00 & -- \\
\midrule
GPT-5.6-sol & Codex & 55.00 & 45.00 & 50.00 & 50.00 & 60.00 & 45.00 & 55.00 & 35.00 \\
GLM-5.3-flash & Codex (3-run avg.) & 56.67 & 40.00 & 18.33 & 8.33 & 53.33 & 10.00 & 38.33 & 25.00 \\
GLM-5.3-flash & Claude Code & 60.00 & 50.00 & 20.00 & 15.00 & 35.00 & 10.00 & 35.00 & 25.00 \\
DeepSeek-v4-flash-0731 & Codex (3-run avg.) & 45.00 & 36.67 & 18.33 & 11.67 & 46.67 & 16.67 & 38.33 & 23.33 \\
DeepSeek-v4-flash-0731 & Claude Code & 35.00 & 20.00 & 25.00 & 25.00 & 30.00 & 15.00 & 35.00 & 25.00 \\
DeepSeek-v4-flash-0731 & DeepSeek Harness & 60.00 & 35.00 & 5.00 & 0.00 & 50.00 & 15.00 & 40.00 & 15.00 \\
DeepSeek-v4-pro-0813 & DeepSeek Harness & 45.00 & 25.00 & 10.00 & 10.00 & 35.00 & 15.00 & 45.00 & 30.00 \\
\bottomrule
\end{tabular}
}
\end{table*}

\subsection{Results by Mutation Count and Variant Subset}
\label{app:variant-results}

Table~\ref{tab:appendix-variant-only-recovery} reports aggregate recovery after excluding the Original instance from every included task family. The resulting variant-only subsets contain 72 Core-set and 496 Full-set instances. For Codex with GPT-5.6-sol, removing the Original instances reduces $\mathrm{SA}(\mathcal P)$ from $35.00\%$ to $30.56\%$ on the Core-set and from $16.02\%$ to $14.11\%$ on the Full-set. The corresponding strict $\mathrm{SA}(\mathcal M)$ values decrease from $13.75\%$ to $9.72\%$ and from $7.81\%$ to $6.05\%$, respectively. The other standard discovery configurations achieve at most $4.17\%$ phenomenal recovery and $1.39\%$ mechanism recovery on the Core-set variants; Codex with GLM-5.3-flash reaches $3.23\%$ and $0.81\%$ on the Full-set variants.

Tables~\ref{tab:appendix-core-mutation-recovery} and \ref{tab:appendix-full-mutation-recovery} further stratify recovery by the number of applied mutations. For Codex with GPT-5.6-sol, Core-set $\mathrm{SA}(\mathcal P)$ declines from $75.00\%$ on Original instances to $75.00\%$, $31.25\%$, $25.00\%$, $6.25\%$, and $0.00\%$ with one through five mutations. Strict $\mathrm{SA}(\mathcal M)$ declines more sharply, from $50.00\%$ on Original instances to $37.50\%$ with one mutation, $6.25\%$ with two mutations, and zero with three or more. The Full-set shows the same broad pattern: phenomenal recovery falls from $75.00\%$ on Original instances to zero at five mutations, while mechanism recovery falls from $62.50\%$ to zero by four mutations.

For the smaller models, most successful recoveries are concentrated among Original and single-mutation instances. On the Core-set, none of these standard discovery configurations recovers a complete mechanism with three or more mutations, and only isolated phenomenal-law successes remain beyond two mutations. The Full-set GLM-5.3-flash configuration similarly reaches only $0.63\%$ phenomenal recovery at three mutations and no phenomenal or mechanism recovery at four or five mutations.

\begin{table*}[t]
    \centering
    \caption{Variant-only symbolic-recovery results after excluding Original instances. Codex with GLM-5.3-flash and Codex with DeepSeek-v4-flash-0731 use three runs only on the Core-set. A dash denotes a metric or set that is not evaluated for the corresponding configuration.}
    \label{tab:appendix-variant-only-recovery}
    \resizeTabular{\begin{tabular}{llcccc}
\toprule
\multirow{2}{*}{\textbf{Base model}} & \multirow{2}{*}{\textbf{Agent}}
& \multicolumn{2}{c}{\textbf{Core-set variants}}
& \multicolumn{2}{c}{\textbf{Full-set variants}} \\
\cmidrule(lr){3-4}\cmidrule(lr){5-6}
& & $\mathrm{SA}(\mathcal P)$ & $\mathrm{SA}(\mathcal M)$ & $\mathrm{SA}(\mathcal P)$ & $\mathrm{SA}(\mathcal M)$ \\
\midrule
No LLM & PySR & 0.00 & -- & 0.00 & -- \\
DeepSeek-v4-flash-0731 & PySR + Direct-Ask & -- & 0.00 & -- & 0.00 \\
\midrule
GPT-5.6-sol & Codex & 30.56 & 9.72 & 14.11 & 6.05 \\
GLM-5.3-flash & Codex (3-run avg. / 1 run) & 2.78 & 0.46 & 3.23 & 0.81 \\
GLM-5.3-flash & Claude Code & 4.17 & 1.39 & -- & -- \\
DeepSeek-v4-flash-0731 & Codex (3-run avg.) & 3.24 & 0.46 & -- & -- \\
DeepSeek-v4-flash-0731 & Claude Code & 2.78 & 1.39 & -- & -- \\
DeepSeek-v4-flash-0731 & DeepSeek Harness & 1.39 & 0.00 & -- & -- \\
DeepSeek-v4-pro-0813 & DeepSeek Harness & 4.17 & 0.00 & -- & -- \\
\bottomrule
\end{tabular}
}
\end{table*}

\begin{table*}[t]
    \centering
    \caption{Core-set symbolic recovery stratified by the number of applied mutations. Mutation count zero denotes Original instances.}
    \label{tab:appendix-core-mutation-recovery}
    \resizeTabular{\begin{tabular}{llcccccccccccc}
\toprule
\multirow{2}{*}{\textbf{Base model}} & \multirow{2}{*}{\textbf{Agent}}
& \multicolumn{6}{c}{\textbf{$\mathrm{SA}(\mathcal P)$ by mutation count}}
& \multicolumn{6}{c}{\textbf{$\mathrm{SA}(\mathcal M)$ by mutation count}} \\
\cmidrule(lr){3-8}\cmidrule(lr){9-14}
& & 0 & 1 & 2 & 3 & 4 & 5 & 0 & 1 & 2 & 3 & 4 & 5 \\
\midrule
No LLM & PySR & 0.00 & 0.00 & 0.00 & 0.00 & 0.00 & 0.00 & -- & -- & -- & -- & -- & -- \\
DeepSeek-v4-flash-0731 & PySR + Direct-Ask & -- & -- & -- & -- & -- & -- & 12.50 & 0.00 & 0.00 & 0.00 & 0.00 & 0.00 \\
\midrule
GPT-5.6-sol & Codex & 75.00 & 75.00 & 31.25 & 25.00 & 6.25 & 0.00 & 50.00 & 37.50 & 6.25 & 0.00 & 0.00 & 0.00 \\
GLM-5.3-flash & Codex (3-run avg.) & 37.50 & 8.33 & 0.00 & 4.17 & 0.00 & 0.00 & 25.00 & 2.08 & 0.00 & 0.00 & 0.00 & 0.00 \\
GLM-5.3-flash & Claude Code & 50.00 & 12.50 & 6.25 & 0.00 & 0.00 & 0.00 & 25.00 & 0.00 & 6.25 & 0.00 & 0.00 & 0.00 \\
DeepSeek-v4-flash-0731 & Codex (3-run avg.) & 50.00 & 8.33 & 6.25 & 0.00 & 0.00 & 0.00 & 33.33 & 2.08 & 0.00 & 0.00 & 0.00 & 0.00 \\
DeepSeek-v4-flash-0731 & Claude Code & 37.50 & 12.50 & 0.00 & 0.00 & 0.00 & 0.00 & 25.00 & 6.25 & 0.00 & 0.00 & 0.00 & 0.00 \\
DeepSeek-v4-flash-0731 & DeepSeek Harness & 50.00 & 6.25 & 0.00 & 0.00 & 0.00 & 0.00 & 25.00 & 0.00 & 0.00 & 0.00 & 0.00 & 0.00 \\
DeepSeek-v4-pro-0813 & DeepSeek Harness & 62.50 & 18.75 & 0.00 & 0.00 & 0.00 & 0.00 & 37.50 & 0.00 & 0.00 & 0.00 & 0.00 & 0.00 \\
\bottomrule
\end{tabular}
}
\end{table*}

\begin{table*}[t]
    \centering
    \caption{Full-set symbolic recovery stratified by the number of applied mutations. Mutation count zero denotes Original instances.}
    \label{tab:appendix-full-mutation-recovery}
    \resizeTabular{\begin{tabular}{llcccccccccccc}
\toprule
\multirow{2}{*}{\textbf{Base model}} & \multirow{2}{*}{\textbf{Agent}}
& \multicolumn{6}{c}{\textbf{$\mathrm{SA}(\mathcal P)$ by mutation count}}
& \multicolumn{6}{c}{\textbf{$\mathrm{SA}(\mathcal M)$ by mutation count}} \\
\cmidrule(lr){3-8}\cmidrule(lr){9-14}
& & 0 & 1 & 2 & 3 & 4 & 5 & 0 & 1 & 2 & 3 & 4 & 5 \\
\midrule
No LLM & PySR & 6.25 & 0.00 & 0.00 & 0.00 & 0.00 & 0.00 & -- & -- & -- & -- & -- & -- \\
DeepSeek-v4-flash-0731 & PySR + Direct-Ask & -- & -- & -- & -- & -- & -- & 6.25 & 0.00 & 0.00 & 0.00 & 0.00 & 0.00 \\
\midrule
GPT-5.6-sol & Codex & 75.00 & 40.00 & 15.00 & 8.13 & 1.25 & 0.00 & 62.50 & 25.00 & 5.63 & 0.63 & 0.00 & 0.00 \\
GLM-5.3-flash & Codex & 37.50 & 13.75 & 2.50 & 0.63 & 0.00 & 0.00 & 31.25 & 5.00 & 0.00 & 0.00 & 0.00 & 0.00 \\
\bottomrule
\end{tabular}
}
\end{table*}

\subsection{Statistical Comparisons}
\label{app:statistical-comparisons}

\paragraph{Base-model comparisons.} GPT-5.6-sol significantly outperforms both flash models across phenomenal recovery, mechanism recovery, and OOD numerical accuracy.  For each comparison, we first average the flash model's binary outcome within each task and then apply a paired Wilcoxon test over the 80 tasks \citep{Wilcoxon1945}.  After Holm correction across the six model--metric comparisons \citep{Holm1979}, GPT-5.6-sol exceeds GLM-5.3-flash by $28.75$ percentage points in $\mathrm{SA}(\mathcal P)$, $10.83$ points in $\mathrm{SA}(\mathcal M)$, and $22.92$ points in OOD $\mathrm{Acc}_{0.1}$, with adjusted $p<.001$, $p=.008$, and $p<.001$, respectively.  Its corresponding advantages over DeepSeek-v4-flash-0731 are $27.08$, $10.00$, and $21.67$ points, with the same adjusted significance levels.  Thus, the performance advantage of GPT-5.6-sol is present in all three evaluated outcomes.

\paragraph{Base-model and harness comparisons.} With Codex fixed, paired Wilcoxon comparisons between DeepSeek-v4-flash-0731 and GLM-5.3-flash detect no difference in phenomenal recovery ($p=.234$), mechanism recovery ($p=.414$), or OOD numerical accuracy ($p=.662$).  Exact McNemar comparisons under Claude Code \citep{McNemar1947} give the same conclusion ($p=.688$, $1.000$, and $.607$).  Holding DeepSeek-v4-flash-0731 fixed and comparing Codex, Claude Code, and DeepSeek Harness on the same 80 tasks, Cochran's $Q$ test \citep{Cochran1950} is nonsignificant for all three metrics ($p=.867$, $.607$, and $.264$).  For GLM-5.3-flash, paired Wilcoxon comparisons between Codex and Claude Code are likewise nonsignificant ($p=.152$, $.655$, and $.075$).  Accordingly, the present comparisons distinguish GPT-5.6-sol from the flash models, but do not establish a performance difference between the two flash models or among the evaluated harnesses.

\paragraph{Domain comparisons.} We find no statistically significant performance difference among physics, chemistry, biology, and materials science.  We summarize each Core-set task by its mean outcome across the 11 standard-discovery runs and compare the resulting four groups of 20 tasks using the Kruskal--Wallis test \citep{KruskalWallis1952}.  The domain effect is nonsignificant for $\mathrm{SA}(\mathcal P)$ ($p=.363$), strict $\mathrm{SA}(\mathcal M)$ ($p=.092$), and OOD $\mathrm{Acc}_{0.1}$ ($p=.149$).  Although physics has the highest descriptive mean on all three metrics, the current evidence does not support a general domain-level performance ordering; the family-level differences reported in Section~\ref{sec:factor-comparisons} instead indicate that difficulty varies substantially within domains.

Table~\ref{tab:appendix-statistical-comparisons} provides the rates and $p$ values for all comparisons discussed above.  For the GPT-5.6-sol comparisons, the displayed $p$ values are Holm-adjusted; the other entries are the corresponding two-sided paired or omnibus tests.

\begin{table*}[t]
    \centering
    \caption{Core-set statistical comparisons across base models, harnesses, and scientific domains.  Rates follow the order of the groups named in each row.  Codex rates for the two flash models are task-level averages over three runs except in the three-harness DeepSeek comparison, which uses the first Codex run alongside the single Claude Code and DeepSeek Harness runs.}
    \label{tab:appendix-statistical-comparisons}
    \resizeTabular{\begin{tabular}{lllcc}
\toprule
Factor & Comparison or grouping & Metric & Rates (\%) & $p$ \\
\midrule
Base model & GPT-5.6-sol vs GLM-5.3-flash (Codex) & $\mathrm{SA}(\mathcal P)$ & 35.00 / 6.25 & $<.001$ \\
& & $\mathrm{SA}(\mathcal M)$ & 13.75 / 2.92 & .008 \\
& & $\mathrm{Acc}_{0.1}^{\mathrm{OOD}}$ & 43.75 / 20.83 & $<.001$ \\
& GPT-5.6-sol vs DeepSeek-v4-flash-0731 (Codex) & $\mathrm{SA}(\mathcal P)$ & 35.00 / 7.92 & $<.001$ \\
& & $\mathrm{SA}(\mathcal M)$ & 13.75 / 3.75 & .008 \\
& & $\mathrm{Acc}_{0.1}^{\mathrm{OOD}}$ & 43.75 / 22.08 & $<.001$ \\
\midrule
Base model & DeepSeek vs GLM (Codex) & $\mathrm{SA}(\mathcal P)$ & 7.92 / 6.25 & .234 \\
& & $\mathrm{SA}(\mathcal M)$ & 3.75 / 2.92 & .414 \\
& & $\mathrm{Acc}_{0.1}^{\mathrm{OOD}}$ & 22.08 / 20.83 & .662 \\
& DeepSeek vs GLM (Claude Code) & $\mathrm{SA}(\mathcal P)$ & 6.25 / 8.75 & .688 \\
& & $\mathrm{SA}(\mathcal M)$ & 3.75 / 3.75 & 1.000 \\
& & $\mathrm{Acc}_{0.1}^{\mathrm{OOD}}$ & 21.25 / 25.00 & .607 \\
\midrule
Harness & Codex / Claude Code / DSH (DeepSeek) & $\mathrm{SA}(\mathcal P)$ & 5.00 / 6.25 / 6.25 & .867 \\
& & $\mathrm{SA}(\mathcal M)$ & 3.75 / 3.75 / 2.50 & .607 \\
& & $\mathrm{Acc}_{0.1}^{\mathrm{OOD}}$ & 23.75 / 21.25 / 16.25 & .264 \\
& Codex vs Claude Code (GLM) & $\mathrm{SA}(\mathcal P)$ & 6.25 / 8.75 & .152 \\
& & $\mathrm{SA}(\mathcal M)$ & 2.92 / 3.75 & .655 \\
& & $\mathrm{Acc}_{0.1}^{\mathrm{OOD}}$ & 20.83 / 25.00 & .075 \\
\midrule
Domain & Physics / chemistry / biology / materials & $\mathrm{SA}(\mathcal P)$ & 15.9 / 10.5 / 6.4 / 6.8 & .363 \\
& & $\mathrm{SA}(\mathcal M)$ & 11.4 / 5.0 / 0.9 / 0.0 & .092 \\
& & $\mathrm{Acc}_{0.1}^{\mathrm{OOD}}$ & 36.8 / 14.5 / 16.4 / 25.0 & .149 \\
\bottomrule
\end{tabular}
}
\end{table*}

\end{document}